\documentclass{article}
\usepackage{amssymb}

\usepackage[final]{corl_2025} %

\usepackage{graphicx}
\usepackage{siunitx}
\usepackage{booktabs}
\usepackage{multirow}
\usepackage{tabularx}
\usepackage{siunitx}
\usepackage{cuted}
\usepackage{caption}
\usepackage{threeparttable}
\usepackage{array}
\usepackage{enumitem}
\usepackage{microtype}
\usepackage[table, dvipsnames]{xcolor}
\usepackage{cuted} %
\usepackage{tikz}
\usepackage{pgfplots}
\usepackage[resetlabels]{multibib}
\newcites{S}{References}

\usepackage{amsmath}
\usepackage{amssymb}

\DeclareMathAlphabet{\mathcal}{OMS}{cmsy}{m}{n}

\def\R{\mathbb{R}}

\def\seThree{\mathfrak{se}(3)}
\def\SEThree{SE(3)}

\newcommand{\mat}[1]{\mathbf{#1}}
\newcommand{\norm}[1]{\left\lVert#1\right\rVert}

\newcommand{\set}[1]{\left\{#1\right\}}
\newcommand{\fset}[2]{\set{#1 \;\middle\vert\; #2}}

\newcommand{\tf}[3]{{^{#1}}{\mat{#2}}{_{#3}}}

\newcommand{\tuple}[1]{\left\langle#1\right\rangle}

\usepackage{xspace}

\newcommand{\ie}{\mbox{i.\,e.}\xspace}

\newcommand{\etal}{\emph{et al.}\xspace}
   
\renewcommand{\[}{\begin{equation}}
\renewcommand{\]}{\end{equation}}

\renewcommand{\baselinestretch}{0.99}

\definecolor{red}{RGB}{255, 0, 0}   %
\definecolor{orange}{RGB}{255, 77, 0}   %
\definecolor{green}{RGB}{0, 128, 0}   %
\definecolor{purple}{RGB}{160, 32, 240}   %
\definecolor{lightblue}{RGB}{52, 155, 235}   %
\definecolor{darkmagenta}{RGB}{204, 51, 139} %

\usepackage[capitalize]{cleveref}

\crefname{figure}{Fig.}{Figs.}
\Crefname{figure}{Figure}{Figures}
\crefname{section}{Sec.}{Secs.}
\Crefname{section}{Section}{Sections}
\Crefname{table}{Table}{Tables}
\crefname{table}{Tab.}{Tabs.}
\crefname{algorithm}{Algo.}{Algos.}
\Crefname{algorithm}{Algorithm}{Algorithms}
\crefname{appendix}{Sec.}{Secs.}
\Crefname{appendix}{Section}{Sections}

\newcolumntype{P}[1]{>{\centering\arraybackslash}p{#1}}

\newcommand{\ours}{ArtiPoint}
\newcommand{\ourdataset}{Arti4D}

\newcommand{\greyrule}{\arrayrulecolor{black!30}\midrule\arrayrulecolor{black}}

\def\smoothingwindow{w_h}
\def\handthreshold{\tau_h}

\usepackage[capitalize]{cleveref}
\crefname{section}{Sec.}{Secs.}
\Crefname{section}{Section}{Sections}
\Crefname{table}{Table}{Tables}
\crefname{table}{Tab.}{Tabs.}

\title{Articulated Object Estimation in the Wild}

\author{
  Abdelrhman Werby\textsuperscript{\footnotesize 1,2*},\
  Martin Büchner\textsuperscript{\footnotesize 1*},\
  Adrian Röfer\textsuperscript{\footnotesize 1*},\
  Chenguang Huang\textsuperscript{\footnotesize 3}, \\ 
  \textbf{Wolfram Burgard\textsuperscript{\footnotesize 3}, \ Abhinav Valada\textsuperscript{\footnotesize 1}}\\
  \textsuperscript{\footnotesize 1}University of Freiburg\ \quad \quad
  \textsuperscript{\footnotesize 2}University of Stuttgart\ \quad \quad
  \textsuperscript{\footnotesize 3}University of Technology Nuremberg\\
}

\begin{document}
\maketitle

\begingroup
  \renewcommand\thefootnote{*}
  \footnotetext{These three authors contributed equally to this work. Work done while at University of Freiburg.}
\endgroup
\vspace{-0.2em}
\begin{abstract}
    Understanding the 3D motion of articulated objects is essential in robotic scene understanding, mobile manipulation, and motion planning. Prior methods for articulation estimation have primarily focused on controlled settings, assuming either fixed camera viewpoints or direct observations of various object states, which tend to fail in more realistic unconstrained environments. In contrast, humans effortlessly infer articulation by watching others manipulate objects. Inspired by this, we introduce \textit{\ours{}}, a novel estimation framework that can infer articulated object models under dynamic camera motion and partial observability. By combining deep point tracking with a factor graph optimization framework, \textit{\ours{}} robustly estimates articulated part trajectories and articulation axes directly from raw RGB-D videos. To foster future research in this domain, we introduce \textit{\ourdataset{}}, the first ego-centric in-the-wild dataset that captures articulated object interactions at a scene level, accompanied by articulation labels and ground-truth camera poses. We benchmark \textit{\ours{}} against a range of classical and learning-based baselines, demonstrating its superior performance on \textit{\ourdataset{}}. We make code and \textit{\ourdataset{}} publicly available at \url{https://artipoint.cs.uni-freiburg.de}.
\end{abstract}

\keywords{Articulated Object Estimation, 3D Scene Understanding, Interactive Perception} 

\begin{figure}[h]
\centering
\includegraphics[width=0.90\linewidth, trim={0 1cm 0 0},clip]{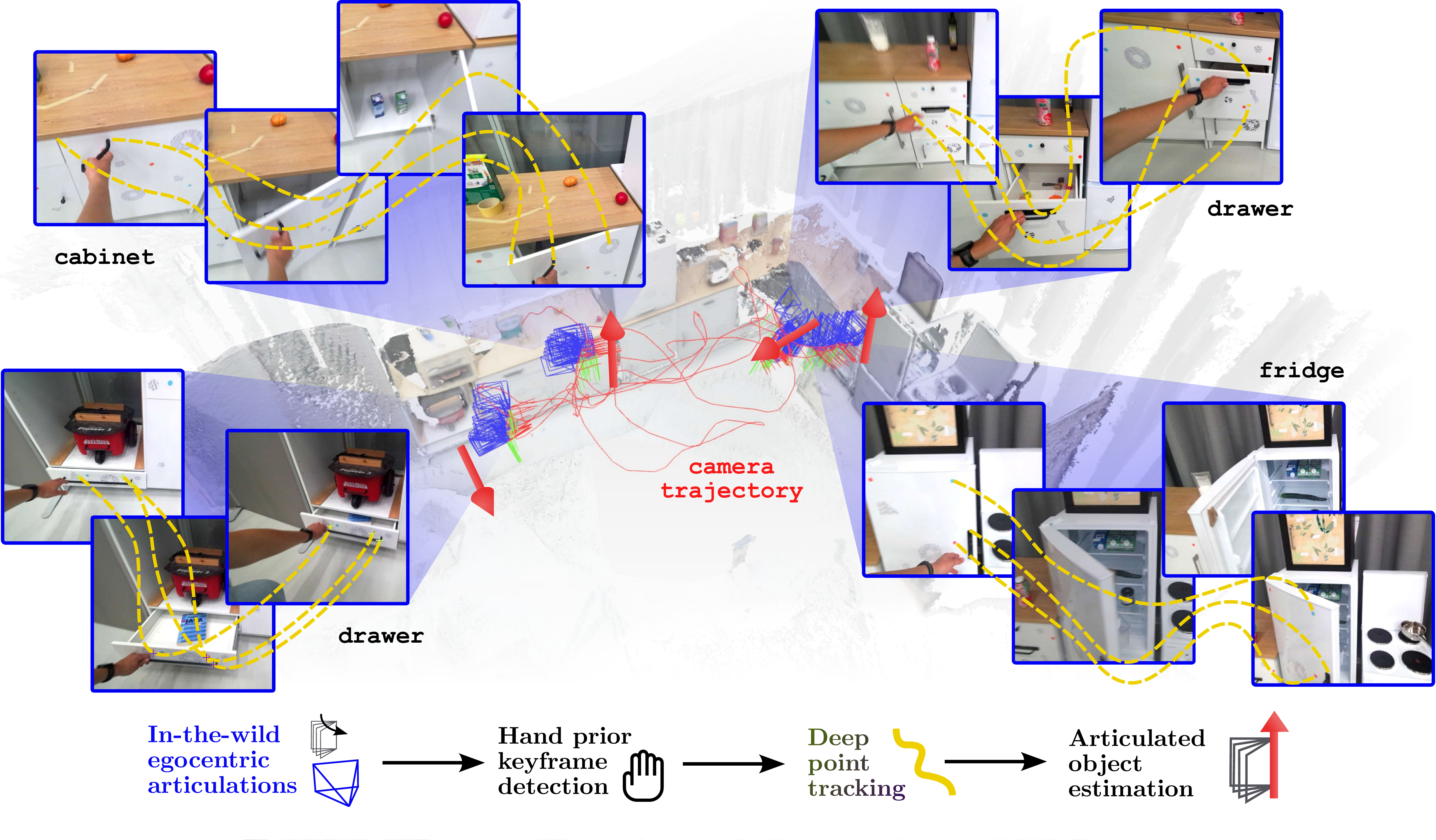}
\caption{
We present \textit{\ours}, our novel approach to articulated object estimation in the wild. \ours{} makes use of deep point tracking and factor graph optimization. 
It does not rely on costly scene-wise optimization or deep models that are prone to overfit. We evaluate our approach on the \ourdataset{} dataset, a novel scene-level articulated object dataset that is recorded with a moving, ego-centric camera, compared to previous benchmarks that are limited to isolated objects and static camera poses.
}
\label{fig:teaser}
\end{figure}

\section{Introduction}
\label{sec:introduction}

Robotic manipulation is concerned with moving external objects to achieve a task-relevant goal. While recent approaches have made tremendous progress in manipulating unconstrained objects~\cite{kim24openvla, chi2024diffusionpolicy, chisari2024learning}, manipulating constrained \emph{articulated} objects remains challenging in model-free settings that rely on affordances~\cite{gupta2024opening,nematollahi2022robot}. On the other hand, model-based approaches have thoroughly investigated manipulating articulated objects, demonstrating both manipulation success and transferability to novel objects~\cite{sturm2008unsupervised, zhang2023flowbot++, honerkamp2023n, buchanan2024online,schmalstieg2023learning}.
However, the aforementioned approaches focus on offline learning of predictive models in curated settings, where only the articulated object itself moves~\cite{liu2023paris, liu2025building}. If we consider deploying a robot in a novel environment, these requirements are not met~\cite{gupta2024opening, momallm24, more25mohammadi}. As a consequence, most approaches for estimating articulated objects and estimating motion models fail when operating \textit{in-the-wild}, \ie, when faced with dynamic camera poses, occlusions, and significant clutter, where objects are not isolated and interactions are less constrained. While humans acquire manipulation skills through observing others interacting with various entities such as articulated objects~\cite{bandura1977social}, endowing robotic systems with those capabilities remains an open problem. Ultimately, this limits the transfer of insights from articulated object estimation~\cite{sturm2008unsupervised, buchanan2024online} to the problem of human-to-robot imitation in everyday robotics.

In this work, we propose to address this limitation by exploiting interaction priors. As humans manipulate environments with their hands, we aim to observe those interactions. First, we extract interaction segments containing object interactions. Secondly, we obtain point trajectories within regions close to the detected hand, which describe both the static and dynamic parts of the observed scene. We leverage recent state-of-the-art models in any-point tracking~\cite{doersch2023tapir, karaev2024cotracker}, which yields point trajectories throughout whole interaction segments, including an estimated visibility score. Third, we lift those point trajectories into 3D using depth measurements and compensate for camera motion using accurate camera odometry. We separate all static 3D trajectories with a negligible length from dynamic trajectories, which potentially represent the motion of dynamic objects. This enables us to estimate the motion of the moving part triggered through hand interaction. Finally, given the acquired point trajectories, we estimate the underlying articulation model, classify the joint type (prismatic or revolute), and globally register the model in the observed scene.

As we address the novel task of estimating articulated object motion \textit{in-the-wild}, we introduce Arti4D, the first ego-centric in-the-wild dataset for scene-level articulated object manipulation from human demonstrations. It comprises 45 RGB-D sequences across four diverse scenes, capturing 414 human-object interactions under dynamic camera motion, occlusions, and uncurated environments ranging from kitchen scenes to robot lab rooms. The dataset was recorded by a human operator holding an RGB-D camera while actively exploring scenes and interacting with articulated objects. Unlike prior datasets, Arti4D includes $\seThree$-labeled ground-truth articulation axes, interaction windows, difficulty annotations, reconstructed scenes, and accurate camera pose ground truth. It offers a challenging benchmark not only for articulated object estimation but also for visual odometry and SLAM research in complex, real-world settings.

Concretely, we make the following contributions:
\begin{enumerate}[topsep=0pt,itemsep=0pt]
    \item We present a novel articulated object estimation framework called \textit{\ours{}} that operates on deep point trajectories, enabling robust deployment in uncontrolled settings.
    \item We introduce \textit{\ourdataset{}}, the first real-world \textit{in-the-wild} dataset of articulated object interactions providing human exploration and ego-centric articulation demonstrations on a scene-level. The dataset includes articulation axes labels and 3D ground truth camera poses.
    \item We compare our method against a set of classical and deep articulation model estimation pipelines and present extensive ablations.
    \item We publish code, data, and model predictions of our method at \url{https://artipoint.cs.uni-freiburg.de}.
\end{enumerate}

\section{Related Work}
\label{sec:related-work}
In the following, we review previous work in articulated object estimation and review recent developments in any-point tracking.

{\parskip=3pt\noindent\textbf{Probabilistic Methods:}}
Early work in articulation model estimation often employed probabilistic formulations to infer the relationships among articulated object parts. Sturm~\etal~\cite{sturm2011probabilistic,sturm2008unsupervised} proposed a probabilistic framework that infers the articulation model from 6D trajectories of object parts. While the initial work used marker-based tracking to obtain these trajectories, subsequent work~\cite{pillai2014learning,katz2014interactive} extended these frameworks to handle sparse (markerless) objects by extracting visual features, such as SURF~\cite{bay2006surf} or by detecting rectangles in dense-depth images~\cite{sturm20103d}.

{\parskip=3pt\noindent\textbf{Deep-Learning-based Methods:}}
Recent advancements in deep learning diversify the articulation estimation methods. By consuming a sequence of depth images with a fixed viewing pose, ScrewNet~\cite{jain2021screwnet} enables category-agnostic articulation motion prediction in an end-to-end manner. A follow-up work, DUST-net~\cite{jain2022distributional} further estimates the uncertainty of the motion to introduce interpretability. While these works take a depth image sequence as input, ANCSH~\cite{li2020category} predicts articulation joint parameters and state for unseen instances using a depth image, assuming prior knowledge of the instance category. Other works utilize neural networks as a part of their pipelines for articulation motion prediction. While Heppert~\textit{et~al.}~\cite{heppert2022category} formulate a two-stage pipeline consisting of a learning-based part tracking stage and a factor graph optimization stage, Buchanan~\textit{et~al.}~\cite{buchanan2024online} combine the learning-based method with interactive perception, leveraging a neural network to propose an initial estimation of the articulation motion and interactive perception for prediction refinement. FormNet~\cite{zeng2021visual} predicted the motion field residual and the part connectedness prior to post-processing the final prediction.

{\parskip=3pt\noindent\textbf{3D Reconstruction of Articulated Objects:}}
Reconstructing articulated objects in 3D is more challenging than reconstructing rigid objects, as it requires modeling the articulation between their constituent parts \cite{heppert2023carto}. Early works~\cite{huang2012occlusion,martin2016integrated} leverage structure-from-motion to reconstruct the articulated object at different configurations, followed by post-processing techniques to estimate joint axes and types. More recent works inspired by Gaussian splatting offer a smooth and continuous representation for complex shapes. For instance, ArtGS~\cite{liu2025building} utilizes 3D Gaussians to jointly optimize for canonical Gaussians and the articulation model. Ditto~\cite{jiang2022ditto} is a neural network-based method for estimating the articulation model and the 3D geometry of an articulated object, given point cloud observations before and after interaction. PARIS~\cite{liu2023paris} takes multiview images at different articulation states and jointly estimates the articulation parameters and the 3D geometry of the object.

{\parskip=3pt\noindent\textbf{Tracking Any Point (TAP):}}
Tracking points is a key problem in various computer vision and robotics tasks, such as 3D reconstruction, video analysis, and object tracking. Early works relied on sparse feature extractors, such as SIFT~\cite{lowe2004distinctive}, SURF~\cite{bay2006surf}, and GFTT~\cite{shi1994good}, as well as optical flow. Tracking any point as a task was introduced in PIP~\cite{harley2022particle}, where the goal was to track pixel locations across entire sequences. TAP-Vid~\cite{doersch2022tap} enhances the problem formulation by providing a benchmark and introducing TAP-Net. Subsequent works \cite{doersch2023tapir, zheng2023pointodyssey, cho2024local, doersch2024bootstap} improved performance while Karaev~\textit{et~al.}~\cite{karaev2024cotracker, karaev2024cotracker3} introduce joint tracking of points to account for their dependencies.

In this work, we address the novel problem of estimating articulations from mobile observations \emph{in-the-wild}. We connect deep any-point tracking~\cite{karaev2024cotracker3} to the estimation approach of Buchanan~\etal~\cite{buchanan2024online} to form a holistic approach to inferring the parameters of multiple articulated objects in extensive, interactive environments.

\section{Approach}
\label{sec:method}
\begin{figure}
    \centering
    \includegraphics[width=1.0\linewidth]{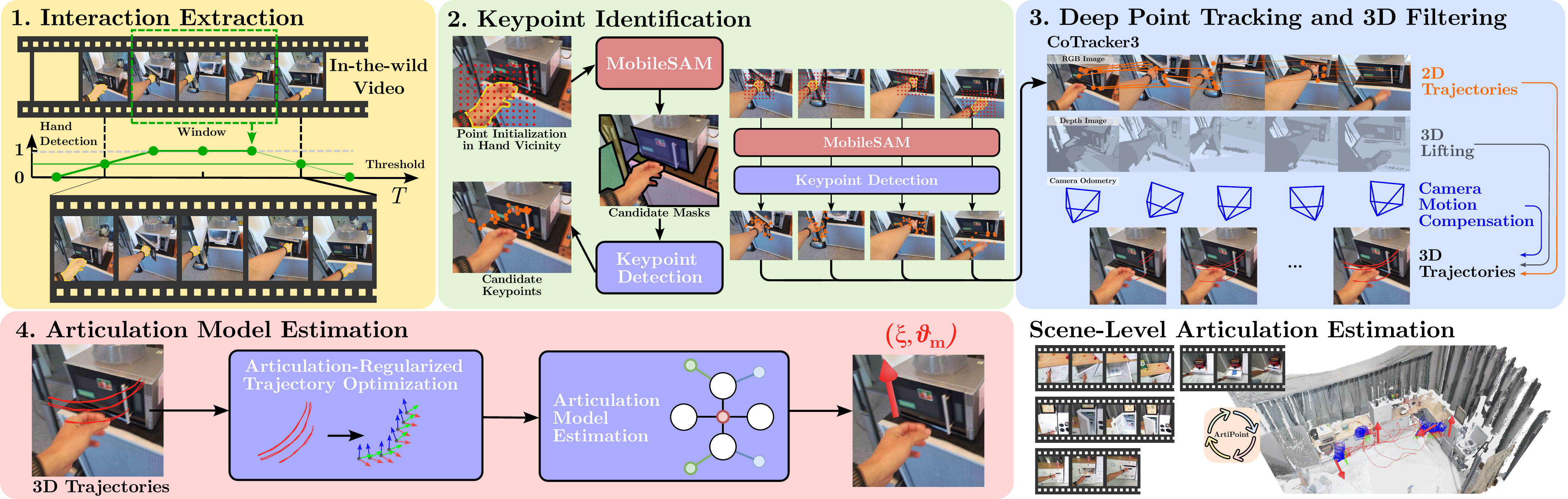}
    \caption{Overview of our method: We take an ego-centric RGB-D video as input and employ hand tracking as a trigger signal to identify interaction segments (top left). We uniformly sample points around the hand masks and prompt a class-agnostic instance segmentation model (MobileSAM~\cite{mobile_sam}), which yields object masks in the immediate vicinity that may be undergoing articulation. Given those masks, we detect stable keypoints (bottom left) that are fed into an any-point tracking model (CoTracker3~\cite{karaev2024cotracker3}) in order to obtain point trajectories throughout each entire articulation segment (top right). Finally, we estimate the underlying articulation model of the object through a factor graph formulation that operates on the obtained point trajectories (bottom right).\looseness=-1}
    \label{fig:overview}
    \vspace{-0.5em}
\end{figure}

The goal of \textit{\ours{}} is to estimate the underlying motion models of articulated objects given posed RGB-D observations of human interactions with such objects. To support this goal, we exploit as a prior that humans typically use their hands to interact with objects. Our approach consists of four stages, which we visualize in \cref{fig:overview}: First, we extract segments that contain hand-object interactions. We then sample 2D keypoints near the hand and track them throughout the segment to capture the object in motion. To obtain 3D motion, we lift the 2D tracks to 3D, compensate for camera motion, filter out static points in 3D, and apply trajectory smoothing in our third stage to mitigate noise. Finally, we input refined 3D tracks into a factor graph framework to jointly estimate the trajectory of the object parts and the parameters of the articulation motion model. 

\subsection{Extraction of Interaction Intervals}
\label{subsec:keyframe_extraction}
Given a scene level ego-centric RGB-D sequence $\set{I_t, D_t}_{t=1}^T$ of a human operating and interacting with articulated objects in the scene, we extract interaction segments $S = \set{(t_{start}^{(n)}, t_{end}^{(n)})}_{n=1}^N$ containing frames of interactions. We use a preexisting hand segmentation model~\cite{camporese2021HandsSeg} to extract a hand mask $H_t$ from each RGB frame $I_t$, and assign a binary label $d_t \in \{0, 1\}$ to the frame based on the hand visibility. To avoid irregular false positives/negatives, we compute the moving average $\bar{d_t}$ for the last $\smoothingwindow$ frames as $\bar{d}_t = \frac{1}{\smoothingwindow} \sum_{i=t-\smoothingwindow}^{t} d_i$.
A frame is identified if the smoothed detection $\bar{d}_t$ exceeds $\handthreshold$, initiating a segment $s_n$ at time $t_{start} = t$. The segment continues as long as $\bar{d}_t \ge \handthreshold$, and terminates at $t_{end} = t$ when $\bar{d}_t$ drops below this threshold. Finally, the extracted segments $S$ are filtered based on their duration, ensuring that $T_{min} \le t_{end}^{(i)} - t_{start}^{(i)} \le T_{max}$, where $T_{min}$ and $T_{max}$ represent the minimum and maximum allowed segment lengths, respectively.

\subsection{Deep Point Tracking}
\label{subsec:point_tracking}
For each obtained segment $s_n \in S$ and hand masks $\set{H_t}_{t=t_{start}^{(n)}}^{t_{end}^{(n)}}$ associated with a frame within the segment, we uniformly sample points around the hand. We use the sampled points as prompts for the class-agnostic instance segmentation model~\cite{mobile_sam} to identify masks of potential articulated objects near the hand for each frame.
Given the resulting $K$ objects' 2D masks $\{O_k^{t}\}_{k=1}^K$ in frame $t$, our goal is to select $F$ query points $q_t \in \R^{F \times 2}$ that lie on the aforementioned masks and allow tracking throughout the entire segment. We achieve this by extracting GFTT keypoints via the lightweight Shi-Tomasi method~\cite{shi1994good} within each object mask in frame $t$. By concatenating all selected query points for a segment $s_n$, we obtain a set of query points $Q_i \in \R^{K \times F \times 2}$ where $K$ represents the number of keyframes selected from the segment $s_n$. Subsequently, we employ CoTracker3~\cite{karaev2024cotracker3} to track keypoints $\mathcal{X}_n \in \R^{T_n \times F \times 2}$ and obtain point visibilites $\mathcal{V}_i \in \{0, 1\}^{T_n \times F}$  of the query points across all $T_n$ frames of each segment $s_n$.

\subsection{3D Track Estimation and Filtering}
\label{subsec:3d_track_estimation}
Given the resulting 2D trajectory of query points $X_n$ for segment $s_n$, first, we lift the trajectory to 3D using the corresponding depth frame $D_t$ and camera intrinsics $\mat{K}$, and we filter out points with invalid depth values. This results in a set of 3D tracks $\mathcal{P}_n \in \R^{T_n \times F \times 3}$ and a corresponding visibility mask $\mathcal{V}_n \in \{0, 1\}^{T_n \times F}$ indicating the validity of each 3D point at each time step. Second, to compensate for camera motion and obtain 3D trajectories in global coordinates, we transform all the 3D point tracks within each segment to the global frame, assuming accurate camera odometry. Third, since some points may lie on static objects, we filter out these static points by computing the positional variance of each 3D track within the segment. Tracks with a variance below a certain percentile threshold $\sigma_{static}$ are considered static and are discarded. Fourth, points may undergo occlusion during the interaction, potentially leading to unreliable point tracks. Therefore, we filter out tracks that are occluded for more than a specified percentage $\sigma_{reliable}$ of the frames within each segment.
Lifting the 2D trajectory to 3D introduces high-frequency temporal jitter in the resulting 3D trajectory. We employ trajectory smoothing via minimizing a cost function (\cref{eq:smoothing_cost}) penalizing spikes in acceleration and high velocities while preserving fidelity
\begin{equation}
\label{eq:smoothing_cost}
E(\mathbf{p}) = \sum_{t=1}^{T} v_t ||\mathbf{p}_t - \mathbf{\hat{p}}_t||^2 + \lambda_{vel} ||\mathbf{p}_t - \mathbf{p}_{t-1}||^2 + \lambda_{jerk} ||\mathbf{p}_t - 3\mathbf{p}_{t-1} + 3\mathbf{p}_{t-2} - \mathbf{p}_{t-3}||^2,
\end{equation}
where the first term promotes closeness between smoothed points $\hat{p}_t$ and observed points $p_t$ weighted by the point visibility $v_t$. The second term regularizes significant velocity changes weighted by $\lambda_{vel}$ whereas the third term minimizes sudden changes in acceleration weighted by $\lambda_{jerk}$.

\subsection{Exploiting the Articulation Prior}
\label{subsec:articulation_estimation}

The most recent works examining articulation model estimation~\cite{jain2021screwnet,heppert2022category,buchanan2024online} opt for representing 1-DoF articulations as an articulation parameter $\xi = \tuple{\omega, v} \in \seThree \subseteq \R^6$. Scaling an element $\xi \in \seThree$ with a configurational position $\theta$ is sufficient to represent any rigid-body transformation which can be decomposed into a proportional linear translation and rotation. Such a transformation $\xi\theta$ can be converted to the equivalent, non-linear $\SEThree$ rigid-body transform using the \emph{exponential map} $\text{exp} : \seThree \rightarrow \SEThree$, while the inverse of the exponential map is referred to as the \emph{log map} $\text{log} : \SEThree \rightarrow \seThree$~\cite{barfoot2024state}. The advantage of this representation is its ability to represent diverse articulations, \ie prismatic, revolute, and screw joints, while also being differentiable which makes them suitable for optimization. 

In this work, we employ the factor graph formulation of Buchanan \etal~\cite{buchanan2024online} to estimate the articulation model which best fits an extracted point trajectory. However, this estimator operates on a sequence of pose observations between two moving parts, not directly on points of these parts. Thus, we extract poses as follows: Given our extracted points $\mathcal{P}_n$ and visibility masks $\mathcal{V}_n$, we interpret these as observation $\mathcal{Z}$ as
\[
    \mathcal{Z} = \set{P_m = \fset{\tuple{p^t_f, p^{t+\epsilon}_f}}{\mathcal{V}^{t}_{n,f} \wedge \mathcal{V}^{t+\epsilon}_{n,f}}}_{m=1}^M,
\]
where $p^t_f, p^{t+\epsilon}_f$ are two observations of point $f$ at times $t$ and $t+\epsilon$ taken from $\mathcal{P}_n$. Extracting pose trajectories from these observations is equivalent to identifying the transformations $\tf{t}{T}{t+\epsilon,m}$ minimizing
\[
\label{eq:rel_traj_opt}
    \tf{t}{T^*}{t+\epsilon,m} = \min_{\tf{t}{T}{t+\epsilon,m}} \sum_{j=1}^{|P_m|} \norm{p^{t+\epsilon}_j - \tf{t}{T}{t+\epsilon,m} \cdot p^t_j},
\]
where $\tuple{p^{t}_j, p^{t+\epsilon}_j} \in P_m$. These local transformations can be integrated to yield poses $\tf{W}{T}{m}$ in the global frame of reference as by recursion: We define $\tf{W}{T}{0} = \tf{W}{T}{A}$ and build on this root with $\tf{W}{T}{m} = \tf{W}{T}{m-1} \cdot \tf{m-1}{T}{m}$
where $\tf{W}{T}{A}$ describes a global offset of the local trajectory.  This local offset can reasonably be obtained as the mean of $\mathcal{P}_n^1$. In our approach, we propose exploiting the context of our problem, an extend the estimation in ~\cref{eq:rel_traj_opt} to be \emph{articulation-regularized} across the entire motion. We do so, by introducing a shared base twist $\hat{\xi}$ for the entire path as 
\[
    \label{eq:articulation_regularized_opt}
    \hat{\xi}^*, \theta^*_1, \ldots, \theta^*_M = \min_{\hat{\xi}, \theta_1, \ldots, \theta_M} \sum_{m=1}^M \sum_{j=1}^{|P_m|} \norm{p^{t+\epsilon}_j - \exp(\hat{\xi}\theta_m) \cdot p^t_j},
\]
where $\tf{t}{T}{t+\epsilon}$ can finally be extracted as $\tf{t}{T}{t+\epsilon,m} = \exp(\hat{\xi}^*\theta^*_m)$. As this formulation can be interpreted as a factor graph, we use GTSAM~\cite{gtsam} to solve for it.

Finally, we extract the relative poses required by the articulation estimator~\cite{buchanan2024online}, as $\tf{A}{T}{m} = \tf{W}{T}{A}^{-1} \cdot \tf{W}{T}{m}$ which is similar to the technique described in~\cite{buchanan2024online}, where the authors use the initial frame of the articulation as the static frame. From this data the estimator then determines $\xi^*_n$ for segment $s_n$.

\section{Arti4D Dataset}
\label{sec:dataset}
We present the \textit{\ourdataset{}} dataset, which, to the best of our knowledge, is the first ego-centric \textit{in-the-wild} human demonstration dataset capturing scene-level articulated object manipulation. With this dataset, we aim to create a foundation for studying the problem of \emph{circumstantial articulated object estimation from human demonstrations}. In this problem, a human collects videos of interactions with multiple objects. The aim is to detect all interactions and estimate the articulation parameters of all manipulated objects. \ourdataset{} consists of 45 egocentric RGB-D sequences across four distinct scenes (\texttt{RR080}, \texttt{DR080}, \texttt{RH078}, and \texttt{RH201}), featuring 414 human-object interactions recorded \emph{in-the-wild}.

Unlike previous datasets such as \textit{PARIS}~\cite{liu2023paris}, \textit{DTA-Multi}~\cite{weng2024neural}, or \textit{ArtGS-Multi}~\cite{liu2025building}, the in-the-wild character presents novel challenges: the camera poses are dynamically changing as the scene is explored, the articulated objects are partially occluded throughout the interaction, and the objects are not isolated, \ie, are an integrated part of the larger environment. We present exemplary interactions and the associated camera trajectory in \cref{fig:teaser}. In this manner, our dataset markedly differentiates itself from existing datasets in the domains of articulated object estimation~\cite{liu2023paris, xiang2020sapien, martin2019rbo} and 3D scene understanding~\cite{rotondi2025fungraph, zhang2025open, kassab2025openlex3d, delitzas2024scenefun3d, werby2024hovsg}.

In tandem with the sequences, our dataset provides ground-truth axis labels for all articulations in the scenes as $\seThree$ points, ground-truth temporal interaction segments, as well as difficulty ratings of all interactions ranked as either \texttt{EASY} or \texttt{HARD}. This is based on the level of hand visibility throughout each interaction, whether an object exhibits a reasonable number of depth measurements, whether the articulating hand is fully retrieved in between articulations, and whether large extents of the object are occluded. Furthermore, we provide cm-accurate ground-truth camera poses, obtained via external tracking, to ease the task of estimating articulations from dynamic camera observations. 
In addition, we believe that \ourdataset{} constitutes a challenging benchmark for the tasks of visual odometry and simultaneous localization and mapping. Given that the articulations cover significant extents of the field of view, finding stable correspondences becomes difficult~\cite{palazzolo2019iros}, ultimately complicating odometry estimation. 

\section{Experimental Results}
\label{sec:experiments}

In the following, we present our experimental findings by comparing against two sets of capable baselines, demonstrating the performance of \ours{} on various splits of \ourdataset{}, and lastly, ablating key components of our proposed pipeline. As introduced in \cref{sec:dataset}, we employ the \ourdataset{} dataset to evaluate the performance of estimating articulation models \textit{in-the-wild}. The reported results are averaged across all object interactions recorded across 45 RGB-D sequences stemming from four diverse environments: \texttt{RR080}, \texttt{DR080}, \texttt{RH078}, and \texttt{RH201}.

{\parskip=3pt\noindent{\textbf{Baselines:}}}
We consider two sets of baselines. The first set consists of the deep learning-based approach Ditto~\cite{hsu2023ditto} and the Gaussian splatting-based method ArtGS~\cite{liu2025building}. Both methods are tailored towards semi-static variations of articulation states of isolated objects. As this is somewhat different from the in-the-wild paradigm introduced as part of the \ourdataset{} dataset, we adapted these methods by masking out hands and selecting a representative number of frames from before, during, and after the interaction (\cref{subsec:keyframe_extraction}) from which to reconstruct the motion model. However, both Ditto and ArtGS are prone to occlusions and variations in camera poses as they aim to identify correspondences among various articulation states. 

Our second set of baselines consists of two factor graph-based articulation estimation pipelines. As Sturm~\textit{et~al.}~\cite{sturm2011probabilistic, sturm2008unsupervised, sturm20103d} only provides a back-end for estimating the articulation model, we employ the \ours{} frontend but fit oriented bounding boxes to all detected objects based on the masks extracted by MobileSAM~\cite{mobile_sam}.
Using the 3D bounding boxes of the objects, we derive part poses across all frames and feed these into the estimation framework of Sturm~\etal~\cite{sturm2011probabilistic}. Lastly, we also employ our front-end pipeline to provide object poses $\tf{W}{T}{A}$ from any-point tracking for Sturm~\etal~\cite{sturm2011probabilistic}.

{\parskip=3pt\noindent{\textbf{Metrics}}}: In order to evaluate estimated articulation models, we match predicted interaction segments against ground truth segments under an $IoU > 0.5$. Given this matching, we employ the following metrics to quantify articulation estimation performance: 1. The angular error $\theta_{err} = \arccos(|\mathbf{\hat{a}} \cdot \mathbf{a}|)$ evaluated between unit-length ground truth and predicted axes, and 2., the Euclidean distance $d_{L2}$ between predicted and ground truth axes (revolute joints only), and the accuracy of predicted joint types.

{\parskip=3pt\noindent{\textbf{Implementation Details:}}}
For our interaction extraction module (\cref{subsec:keyframe_extraction}) we use $w_h=6$, $T_{min}=30$, and $T_{max}=90$. For our 3D track estimation and filtering (\cref{subsec:3d_track_estimation}) we choose a track reliability threshold $\sigma_{reliable}=0.5$, and for the optimization-based smoothing we choose $\lambda_{vel}=0.5$, $\lambda_{jerk}=5$. Our approach requires 20~GB of RAM and 16GB of VRAM.

\subsection{Quantitative Results on Arti4D}
We report an overall comparison on \ourdataset{} in \cref{tab:overall-results}. We note that ArtGS \cite{liu2025building} and Ditto \cite{jiang2022ditto} perform poorly in both estimating and classifying joints, which is primarily due to partial observability throughout interactions. In addition, we observe that the non-isolatedness of objects complicates estimation accuracy in the case of ArtGS and Ditto. Using the estimator proposed by Sturm~\etal~\cite{sturm2011probabilistic} with a bounding box-based front-end leads to worse results compared to ArtGS \cite{liu2025building} and Ditto \cite{jiang2022ditto}. Furthermore, we find that the estimator of Buchanan~\etal~\cite{buchanan2024online} outperforms Sturm~\etal~\cite{sturm2011probabilistic}, making use of a bounding box-based frontend approach as proposed in~\cite{sturm2011learning}. 
Ultimately, we find that the pose trajectories obtained using the independent-transform estimator operating on a frame-pair basis (\cref{eq:rel_traj_opt}) yield higher rotational and translational errors compared to ArtiPoint, Thus, the proposed articulation-regularized estimator (\cref{eq:articulation_regularized_opt}) increases articulation model estimation accuracy. We note that our trajectory estimator also benefits the estimation approach of Sturm~\etal. We also study the performance of the different approaches across easy and difficult interactions in \cref{tab:difficulty-results}. We find that \ours{} performs best in both cases in parameter estimation.

\begin{table}
    \centering
    \scriptsize
    \setlength{\tabcolsep}{8pt}
    \caption{Overall comparison of classical and deep articulation estimation methods on the \textit{\ourdataset{}} dataset. We report the axis-angle errors $\theta_{err}$, the positional errors $d_{L2}$ of revolute joints, and the accuracy of joint type detection. We note that \emph{ArtiPoint} achieves the lowest error in estimating articulation parameters.}
    \label{tab:quantitative_results}
\begin{tabularx}{0.9\linewidth}{l|cccccc}
\toprule
  \multirow{2}{*}{Method} & \multicolumn{2}{c}{Prismatic joints} & \multicolumn{2}{c}{Revolute joints} & \multicolumn{2}{r}{Type accuracy [\%]} \\
 & $\theta_{err}$[deg] & $d_{L2}$[m] & $\theta_{err}$[deg] & $d_{L2}$[m] & Prismatic & Revolute \\
\midrule
ArtGS & 52.29 & -- & 56.82 & 0.25 & \textbf{1.00} & 0.00 \\
Ditto & 55.03 & -- & 60.89 & 0.29 & 0.00 & \textbf{1.00} \\
\greyrule
Sturm~\etal~\cite{sturm2011probabilistic} w/ bbox & 63.15 & -- & 57.98 & 1.34 & 0.00 & 1.00 \\ %
Buchanan ~\cite{buchanan2024online} w/ bbox &  49.67 & -- & 58.82 & 0.22 & 0.00 & 1.00 \\ %
ArtiPoint w/ Sturm~\etal~\cite{sturm2011probabilistic} & 26.85 & -- & 18.32 & 1.40 & 0.70 & 0.96 \\ %
ArtiPoint w/ indep. transforms &  15.60 & -- & 18.61 & 0.15 & 0.65 & 0.96 \\ %
 \greyrule
 ArtiPoint (ours) & \textbf{14.54} & -- & \textbf{17.14} & \textbf{0.07} & 0.68 & 0.98\\ 
\bottomrule
\end{tabularx}
    \label{tab:overall-results}
    \vspace{-0.5cm}
\end{table}

\begin{table}[]
    \centering
    \scriptsize
    \setlength{\tabcolsep}{6pt}
    \caption{Difficulty-level comparison of estimation methods on \textit{\ourdataset{}}.}
    \begin{tabular}{ll|cccccc}
    \toprule
     &  & \multicolumn{2}{c}{Prismatic joints} & \multicolumn{2}{c}{Revolute joints} & \multicolumn{2}{c}{Type accuracy [\%]} \\
    Diff.  & Method & $\theta_{err}$[deg] & $d_{L2}$[m] & $\theta_{err}$[deg] & $d_{L2}$[m] & Prismatic & Revolute \\
    \midrule
    \multirow[t]{4}{*}{\texttt{EASY}} & ArtGS & 53.65 & - & 54.06 & 0.30 & \textbf{1.00} & 0.00 \\
     & Ditto & 54.46 & - & 51.83 & 0.37 & 0.00 & \textbf{1.00} \\
     & ArtiPoint w/ Sturm~\etal~\cite{sturm2011probabilistic} & 20.24 & - & 13.11 & 0.96 & 0.82 & 0.95 \\
    \cmidrule{2-8}
     & ArtiPoint  & \textbf{12.63} & - & \textbf{11.66} & \textbf{0.05} & 0.75 & 0.95 \\
    
    \midrule
    \multirow[t]{4}{*}{\texttt{HARD}} & ArtGS & 50.12 & - & 59.48 & 0.20 & \textbf{1.00} & 0.00 \\
     & Ditto & 56.09 & - & 67.11 & 0.23 & 0.00 & \textbf{1.00} \\
     & ArtiPoint w/ Sturm~\etal~\cite{sturm2011probabilistic} & 38.94 & - & 22.10 & 1.80 & 0.48 & 0.96 \\
    \cmidrule{2-8}
    & ArtiPoint & \textbf{18.03} & - & \textbf{21.10} & \textbf{0.87} & 0.54 & 1.00 \\
    \bottomrule
    \end{tabular}
    \label{tab:difficulty-results}
\end{table}

\begin{table}
    \centering
    \scriptsize
    \setlength{\tabcolsep}{6pt}
    \caption{Ablations of key components of our approach on the \textit{\ourdataset{}} dataset. As before, we report the same set of metrics as in \cref{tab:overall-results}.}
    \begin{tabularx}{0.85\linewidth}{l|ccccccc}
        \toprule
          & \multicolumn{2}{c}{Prismatic joints} & \multicolumn{2}{c}{Revolute joints} & \multicolumn{2}{c}{Type Accuracy}~[\unit{\percent}]\\
        & $\theta_{err}$ [\unit{\deg}] & $d_{L2}$ [\unit{\cm}] & $\theta_{err}$ [\unit{\deg}] & $d_{L2}$ [\unit{\cm}] & Prismatic & Revolute  \\
        \midrule
        w/ ORB keypoints & 19.22 & - & 25.61 & 0.10 & 0.55 & 0.92 \\
        w/ keyframe stride 4 & 14.67 & - & 20.48 & 0.08 & 0.67 & 0.94 \\
        w/o trajectory smoothing & 16.24 & - & 19.38 & 0.11 & 0.58 & 0.92 \\
        w/ unreliable tracks  & 14.61 & - & 18.86 & 0.09 & 0.67 & 0.93 \\
        \greyrule
        \ours{} (ours) & \textbf{14.54} & -- & \textbf{17.14} & \textbf{0.07} & 0.68 & 0.98 \\
        \bottomrule
    \end{tabularx}
    \label{tab:ablations}
\end{table}

\subsection{Qualitative Results on Arti4D}
\begin{figure*}[h!]
\centering
\footnotesize
\setlength{\tabcolsep}{0.01cm}%
\begin{tabularx}{\linewidth}{ccc}
\includegraphics[width=0.33\linewidth]{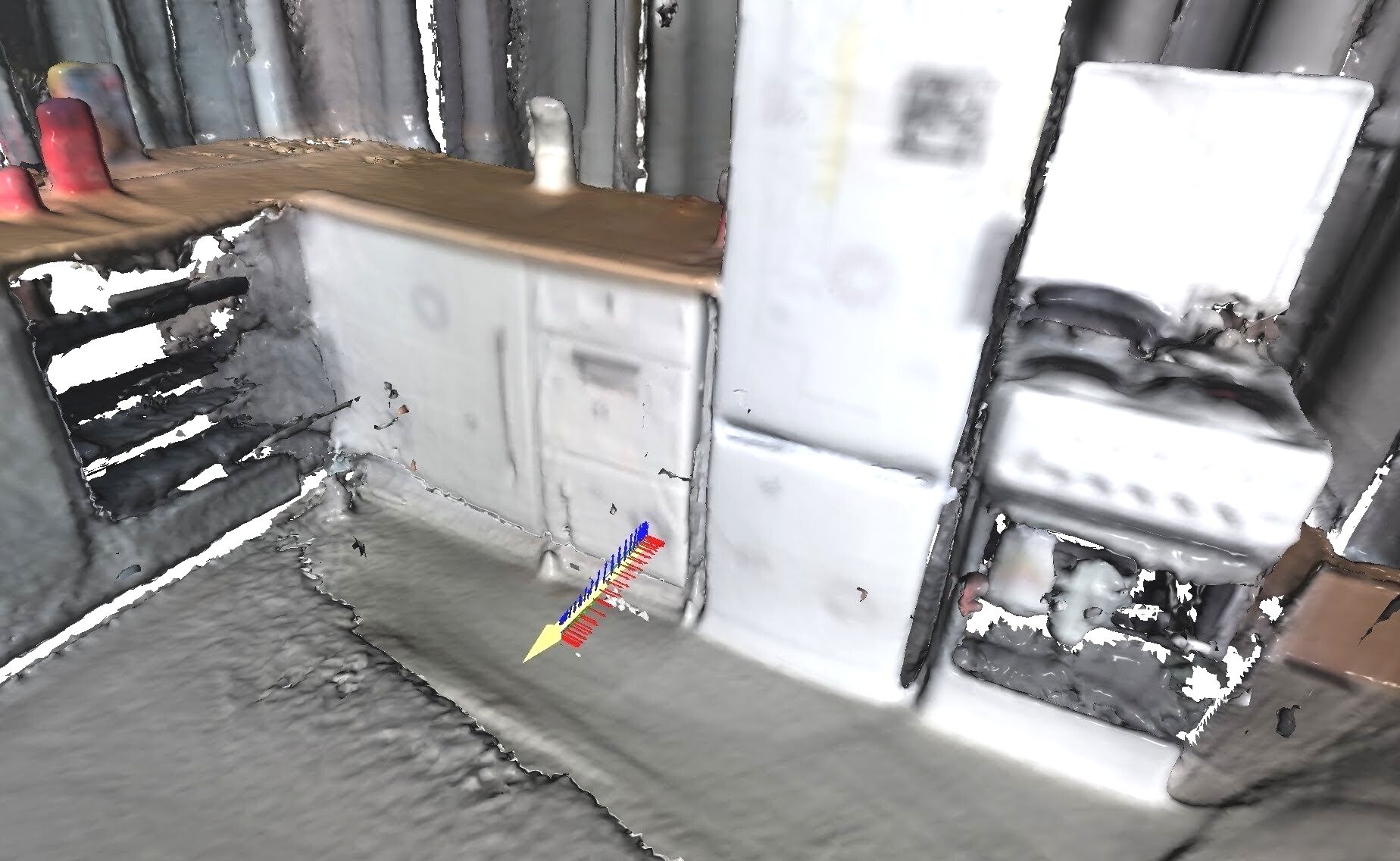} & \includegraphics[width=0.33\linewidth]{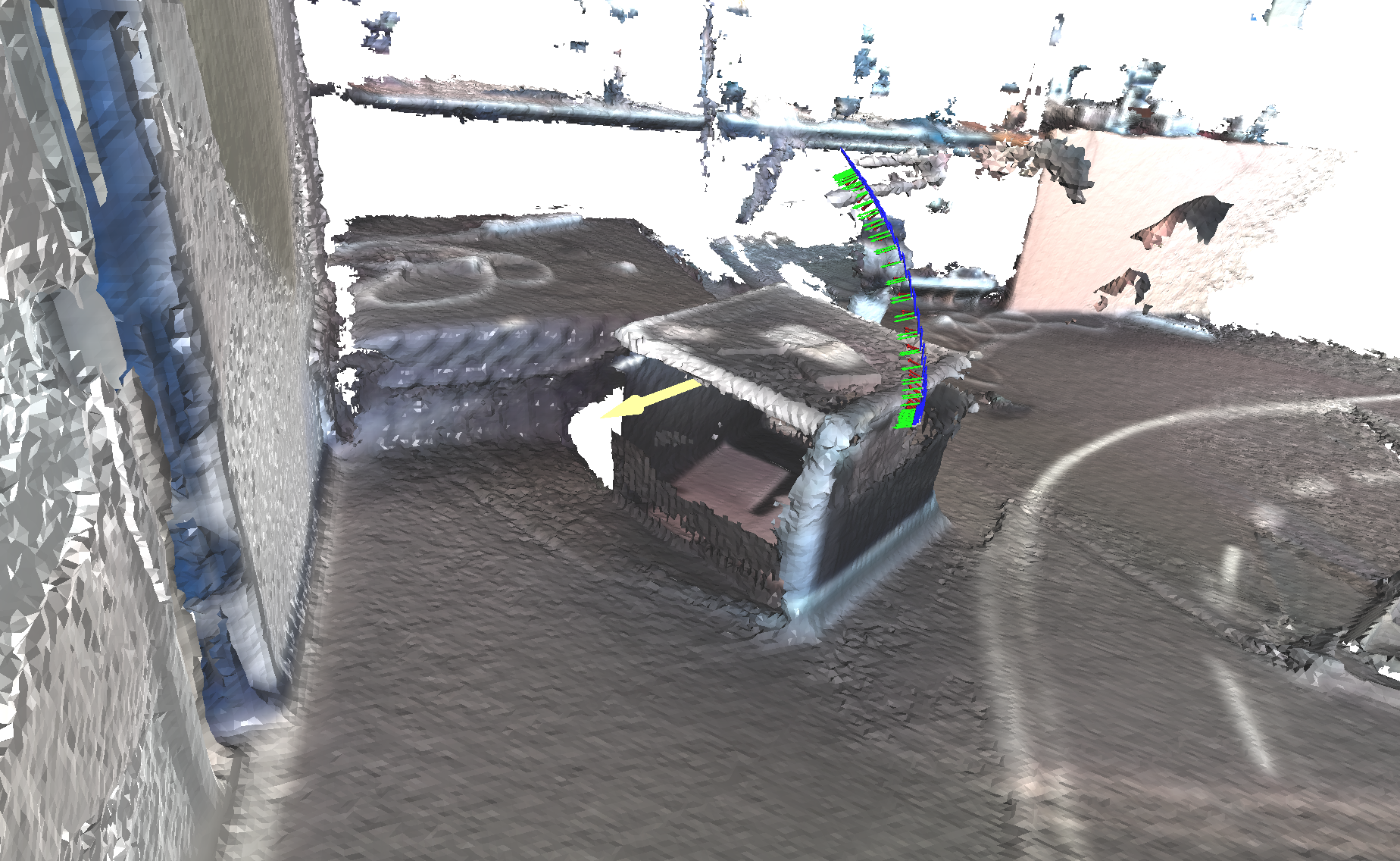} &
\includegraphics[width=0.33\linewidth]{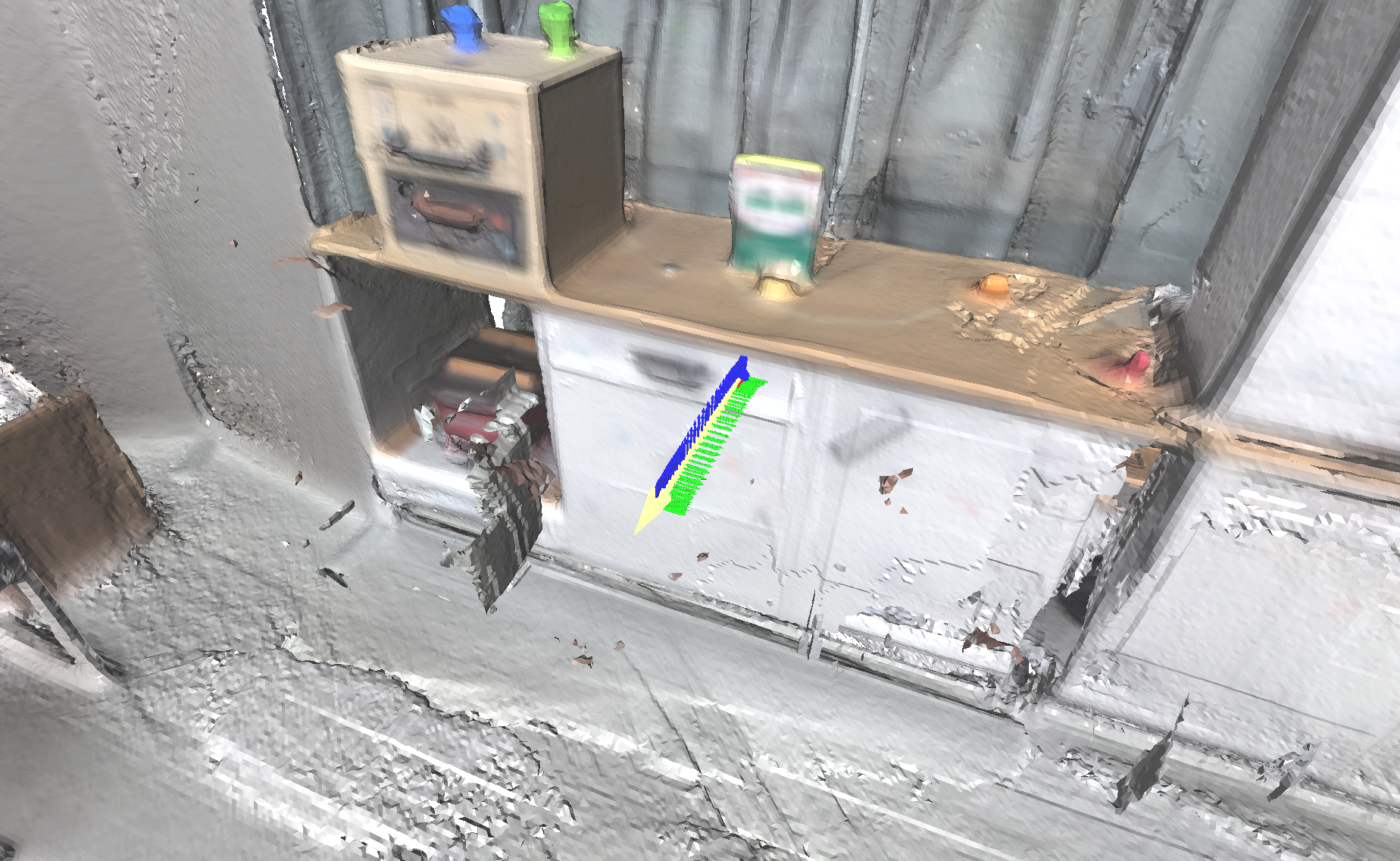} \\
\end{tabularx}
\caption{Qualitative results on \ourdataset{}: Estimated joint axis and pose trajectory of a drawer in a kitchen scene (\texttt{RH201}), a revolute joint of a storage case (\texttt{RH078}), and another drawer in \texttt{RH201}.} 
\label{fig:qualitative}
\vspace*{-0.5cm}
\end{figure*}

In addition to the quantitative comparisons, we depict predictions on two drawers (prismatic) and a revolute joint of a storage case in \cref{fig:qualitative}. The two drawers are observed under a small number of keypoints whereas the case exhibits frequently missing depth measurements. Nonetheless, \ours{} is able to estimate the underlying motion model robustly.

\subsection{Ablation Study}
To further understand the contribution of different components of the \ours{} pipeline, we conducted an ablation study on the \ourdataset{} dataset. From \cref{tab:ablations}, we can conclude that using ORB keypoints \cite{6126544} leads to worse results across all metrics. Increasing the keyframe stride to $4$ in \cref{subsec:point_tracking} reduces the computation overhead but also reduces estimation performance. Furthermore, deactivating the trajectory smoothing impacts the accuracy of both estimated revolute and rotational joints, as well as reducing the overall joint type prediction accuracy. Finally, incorporating unreliable tracks or highly occluded tracks slightly decreases the predicted axis quality, proving the robustness of our proposed estimator.

\section{Conclusion}
\label{sec:conclusion}
We presented \ours{}, a novel framework for estimating articulation motion models from human object interactions \textit{in-the-wild}. Unlike prior works operating in controlled settings involving fixed camera poses, isolated objects, and full observability, we employ deep any-point tracking and factor graph optimization to infer articulation models from ego-centric human demonstrations. As part of our work, we introduced \ourdataset{}, the first real-world, ego-centric articulation demonstration dataset that includes interaction segments, axis labels, and ground truth camera odometry. \ours{} demonstrates robust performance while outperforming both classical and previous deep and object-wise rendering methods in articulated object estimation. To foster further research in articulation understanding in realistic, unconstrained settings, we make code, model predictions, and dataset publicly available.

\newpage
\section{Limitations}
\label{sec:Limitations}
Our proposed \ours{} framework relies on key point tracking, which inherently requires the presence of somewhat distinct features that allow tracking. State-of-the-art any-point tracking methods as discussed in \cref{sec:related-work} show strong robustness even under feature-sparse conditions. However, when objects do not yield dependable depth estimates, due to properties such as color, surface paint, or reflections, tracking their motion becomes particularly challenging. To address this limitation, future work will explore the integration of inpainting methods informed by monocular depth estimation. Additionally, our system is currently limited to relatively simple articulation models that involve only two-body kinematics, making it infeasible to estimate more complex articulations. Another limitation of our approach is the requirement for a trigger signal to temporally segment the RGB-D sequence. In our implementation, this trigger takes the form of human hand detection, which is chosen due to its reliability in indicating articulations. Alternative methods in action recognition, whether in egocentric or third-person views, often lack robustness or require extensive inference times.

\acknowledgments{This work was funded by the BrainLinks-BrainTools center of the University of Freiburg and an academic grant from NVIDIA.}

\bibliography{references}  %

\clearpage
\renewcommand{\baselinestretch}{1}
\setlength{\belowcaptionskip}{0pt}

\setcounter{section}{0}
\setcounter{equation}{0}
\setcounter{figure}{0}
\setcounter{table}{0}
\setcounter{page}{1}
\makeatletter

\renewcommand{\thesection}{S.\arabic{section}}
\renewcommand{\thesubsection}{S.\arabic{section}.\arabic{subsection}}
\renewcommand{\thetable}{S.\arabic{table}}
\renewcommand{\thefigure}{S.\arabic{figure}}

\begin{center}
    {\Large{\bf Articulated Object Estimation in the Wild: \\[0.5em] Supplementary Material}}
\end{center}

In this supplementary material, we provide additional insights on the evaluation protocol and utilized metrics in \cref{sec:metrics-supp}, shed light on the necessity of ground truth camera poses to make ArtiPoint work in real-world deployment in \cref{sec:estimated-poses}, evaluate to what degree ground truth interaction segments benefit the prediction performance in \cref{sec:gt-segments} compared to predicted ones, and provide additional ablations regarding the hand detection as well as the point tracking in \cref{sec:add-ablation-study}. Furthermore, we provide qualitative insights in \cref{sec:qualitative-results-suppl} and in-depth explanations regarding the introduced Arti4D dataset in \cref{sec:arti4d-suppl}.

\section{Experimental Evaluation}

In this section, we present additional details on our experiments, the metrics we employ, and the additional ablation study.

\subsection{Evaluation Protocol \& Metrics}
\label{sec:metrics-supp}

In \cref{sec:experiments}, we report quantitative results of our approach and the baselines. In the following, we detail the prediction to ground truth association procedure as well as the definition of the metrics that we employ to quantify performance.

In order to account for the fact that multiple interactions with the same object instance are likely throughout a single sequence, we match each obtained axis prediction against the corresponding ground truth axes based on the underlying interaction windows. This entails computing the 1-D intersection-over-union (IoU) between all predicted interaction segments and all ground truth interaction segments, as labeled as part of Arti4D. We consider a match whenever an $IoU > 0.5$ is exceeded.

As stated in the main manuscript, our metrics consist of positional and angular error.
We compute the positional error $d_i$ for the placement of an articulation's rotation axis $\hat{\mathbf{a}}$ and the ground truth axis $\mathbf{a}_{gt}$ using the help of supporting points $\hat{\mathbf{p}}, \mathbf{p}_{gt}$ as
\begin{equation}
    d_i = \begin{cases}
        \frac{(\hat{\mathbf{p}}_i - \mathbf{p}_{gt})^\top (\hat{\mathbf{a}}_i \times \mathbf{a}_{gt})}{\norm{\hat{\mathbf{a}}_i \times \mathbf{a}_{gt}}} & \text{if } \norm{\hat{\mathbf{a}}_i \times \mathbf{a}_{gt}} > \epsilon \\
        \norm{(\hat{\mathbf{p}}_i - \mathbf{p}_{gt}) \times \mathbf{a}_{gt}} & \text{else } 
    \end{cases},
\end{equation}
where the first case covers the case in which the axes are not parallel with $\epsilon = 10^{-4}$. If a model does not provide a point on the axis directly, but a twist, we compute $\hat{\mathbf{p}}_i = \frac{\omega_i \times v_i}{\norm{\omega_i}^2}$.
The angular error of a prediction, we compute simply by using the normalized dot-product of the axes

\begin{equation}
    \Theta_{err,i} = \cos^{-1} \left(\frac{\mathbf{a}_{gt}^\top \hat{\mathbf{a}}_i }{\norm{\mathbf{a}_{gt}}\norm{\hat{\mathbf{a}}_i}}\right).
\end{equation}

We report only angular errors for prismatic joints, as the location of the axis does not have any effect on the motion of the parts of the articulated object.

\subsection{Performance Under Estimated Camera Poses}
\label{sec:estimated-poses}

The experimental results reported in the main manuscript rely on ground truth camera odometry from the Arti4D dataset. However, in order to account for direct real-world deployment of our method, we have conducted additional experiments that do not require access to those. To do so, we have evaluated a number of prominent RGB-D SLAM approaches that produce metric map estimates: ORB-SLAM3~\cite{campos2021orb}, Open-VINS~\cite{genenva2020openvins}, MAST3R-SLAM~\cite{murai2024_mast3rslam}, and DROID-SLAM~\cite{teed2021droid}. While all deep learning-based methods provide globally consistent maps, all traditional approaches fail due to loss of camera tracking. Given their non-static character, the interaction segments pose the greatest challenge. As our method requires camera odometry instead of only keyframe-level pose estimates, we employ DROID-SLAM over MAST3R-SLAM.

First, we note that the maps produced by DROID-SLAM are registered against the ground truth point cloud using KISS-Matcher~\cite{lim2025kissmatcher} in order to provide grounds for evaluation, which, in turn, may induce translational and rotational errors. When utilizing DROID-SLAM camera poses, we achieve reasonable results with only slightly increased angular and translational errors for both prismatic and revolute joints. Nonetheless, we observe a slight increase in prismatic type prediction accuracy when employing DROID-SLAM poses.

\begin{table}[h]
    \centering
    \scriptsize
    \setlength{\tabcolsep}{8pt}
    \caption{Comparison of estimated object axis under estimated and ground truth camera poses}
    \begin{tabular}{l|cccccc}
    \toprule
      \multirow{2}{*}{Method} & \multicolumn{2}{c}{Prismatic joints} & \multicolumn{2}{c}{Revolute joints} & \multicolumn{2}{r}{Type accuracy [\%]} \\
      & $\theta_{err}$[deg] & $d_{L2}$[m] & $\theta_{err}$[deg] & $d_{L2}$[m] & Prismatic & Revolute \\
    \midrule
     w/ DROID-SLAM poses & 14.67 & -- & 18.12 & 0.10 & 0.71 & 0.94 \\
     w/ Arti4D odometry & 14.54 & -- & 17.14 & 0.07 & 0.68 & 0.98\\
    \bottomrule
    \end{tabular}
    \label{tab:estimated-camera-poses}
\end{table}

\subsection{Evaluation Using Ground-Truth Interaction Segments}
\label{sec:gt-segments}

In addition to the results relying on predicted interaction segments using the windowed hand detection scheme, we provide an evaluation based on ground truth interaction segments that are labeled as part of Arti4D. In general, we would expect lower angular and translational errors given that the method is not affected by hand occlusions (under large opening angles), motion blur, or non-detected hands. However, we find that our articulation estimation is robust wrt. to the interaction segmentation as we observe smaller angular errors on prismatic objects but larger errors for revolute-jointed objects, thus not allowing a clear interpretation.

\begin{table}[h]
    \centering
    \scriptsize
    \setlength{\tabcolsep}{8pt}
    \caption{Evaluation under ground-truth interaction segments}
    \begin{tabular}{l|cccccc}
    \toprule
      \multirow{2}{*}{Method} & \multicolumn{2}{c}{Prismatic joints} & \multicolumn{2}{c}{Revolute joints} & \multicolumn{2}{r}{Type accuracy [\%]} \\
      & $\theta_{err}$[deg] & $d_{L2}$[m] & $\theta_{err}$[deg] & $d_{L2}$[m] & Prismatic & Revolute \\
    \midrule
     w/ GT segments & \textbf{11.99} & -- & 20.24 & 0.09 & \textbf{0.74} & 0.97 \\
     w/ pred. segments & 14.54 & -- & \textbf{17.14} & \textbf{0.07} & 0.68 & \textbf{0.98} \\
    \bottomrule
    \end{tabular}
    \label{tab:ground-truth-segments}
\end{table}

\subsection{Additional Ablation Study}
\label{sec:add-ablation-study}

In the following sections, we share additional ablations on hyperparameters of ArtiPoint and scene-respective results.

\subsubsection{Ablation on Hand Detection}
\begin{figure}[h]
    \centering
    \includegraphics[width=0.99\linewidth, clip, trim={0.2cm 10.5cm 0cm 0.6cm}]{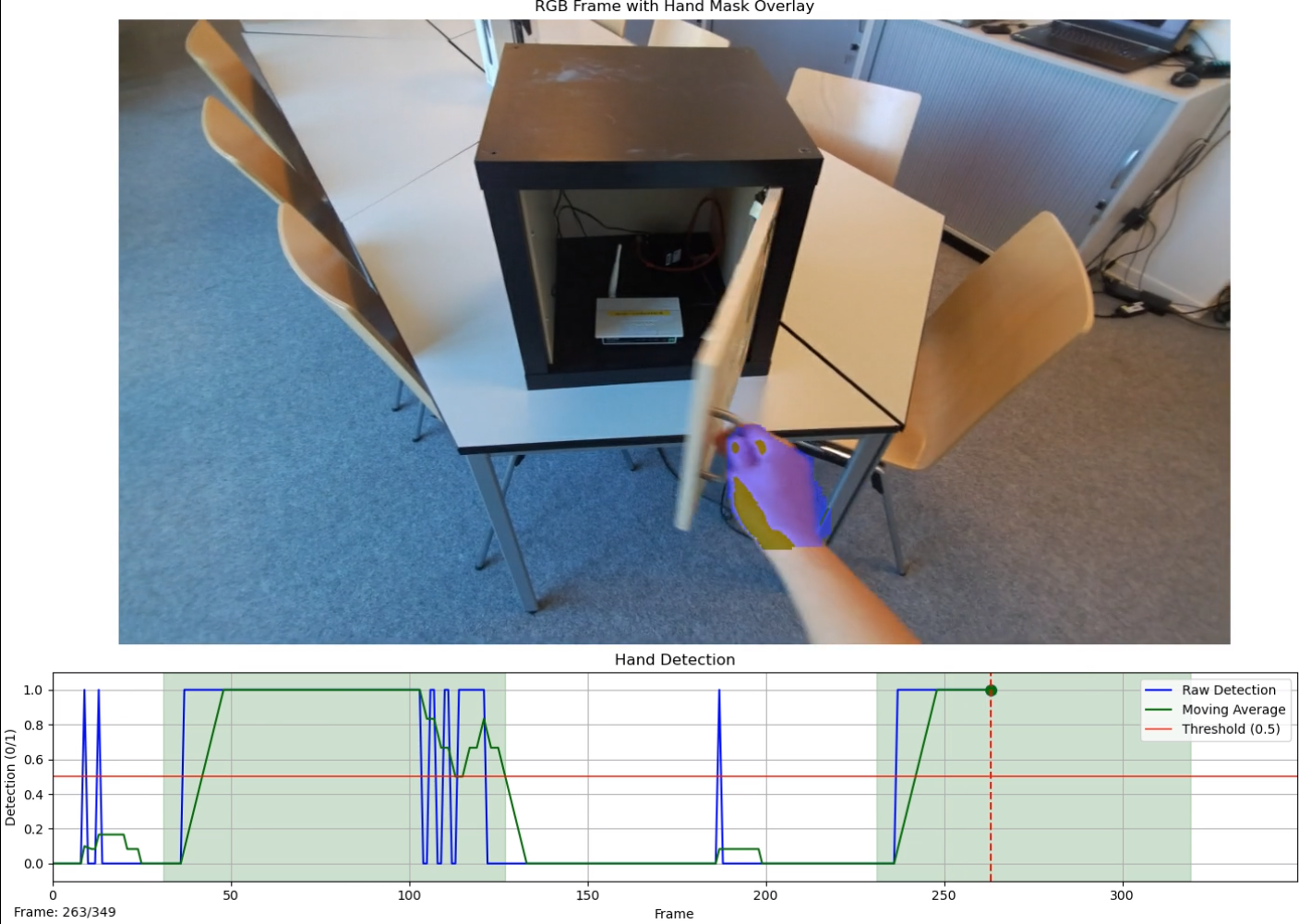}
    \includegraphics[width=0.885\linewidth, clip, trim={0.2cm 0cm 0cm 25.5cm}]{figures/hand-detection.png}
    \caption{Hand detection and interaction extraction: We visualize a live frame of an interaction, including a hand mask marked violet (top). In addition, we visualize the frequency of raw hand detections over time up to the live frame as well as its moving average (bottom). The horizontal red line indicates the threshold at which an interaction segment is created, given the moving average signal. The vertical dashed red line indicates the current frame.}
    \label{fig:hand-detection-viz}
\end{figure}

Extracting the interaction intervals, as described in \cref{subsec:keyframe_extraction} and illustrated in \cref{fig:hand-detection-viz}, is a critical component of the ArtiPoint pipeline as it directly affects the number of articulated objects detected. As such, it requires careful parameter tuning, as small values of $w_h$ or $T_{min}$ increase the number of false positives. To better understand the impact of these parameters on the final results, we conduct a detailed ablation study with its findings presented in \cref{tab:ablation_hand_detection}. Decreasing $T_{min}$ leads to a small degradation in both angular error ($\theta_{err}$) and joint type classification accuracy. Increasing $w_h$ leads to over-smoothing of the raw hand detection signal, causing the segments to contain elongated idle phases at the start and the end, thus inducing unfavorable noise, and resulting in a noticeable degradation in both angular error ($\theta_{err}$) and joint type classification accuracy for both joint types. Furthermore, we also list the effect of not filtering outlier tracks, not using backward tracking, using CoTracker2 over CoTracker3, and performing static trajectory using three-dimensional trajectory data instead of two-dimensional data that is less subject to noisy depth.

\begin{table}[h]
    \centering
    \scriptsize
    \setlength{\tabcolsep}{8pt}
    \caption{Ablations of key parameters for extracting the interaction intervals component \cref{subsec:keyframe_extraction} on the Arti4D dataset. As before, we report the same set of metrics as in \cref{tab:overall-results}. Default values: $T_{max}=90, T_{min}={30}, w_h = 6$. }
    \begin{tabular}{l|cccccc}
    \toprule
      \multirow{2}{*}{Method} & \multicolumn{2}{c}{Prismatic joints} & \multicolumn{2}{c}{Revolute joints} & \multicolumn{2}{r}{Type accuracy [\%]} \\
     & $\theta_{err}$[deg] & $d_{L2}$[m] & $\theta_{err}$[deg] & $d_{L2}$[m] & Prismatic & Revolute \\
    \midrule
     w/o filtering outlier tracks & 13.17 & -- & 22.80 & 0.14 & 0.64 & 0.84 \\
     w/o backward tracking & 13.92 & -- & 20.67 & 0.13 & 0.71 & 0.94 \\
     CoTracker2 & 15.78 & -- & 21.04 & 0.10 & 0.66 & 0.94 \\
     static traj. filter in 3D & 15.18 & -- & 18.23 & 0.10 & 0.65 & 0.92\\ %
     $w_h=12$ & 15.58 & -- & 19.88 & 0.08 & 0.66 & 0.97 \\
     $T_{max}=120$ & 14.57 & -- & 18.13 & 0.07 & 0.68 & 0.96\\
     $T_{min}=15$ & 14.57 & -- & 17.88 & 0.07 & 0.68 & 0.96\\
     \greyrule
     ArtiPoint & 14.54 & -- & 17.14 & 0.07 & 0.68 & 0.98\\
    \bottomrule
    \end{tabular}
    \label{tab:ablation_hand_detection}
\end{table}

\subsubsection{Ablation on Point Tracking}

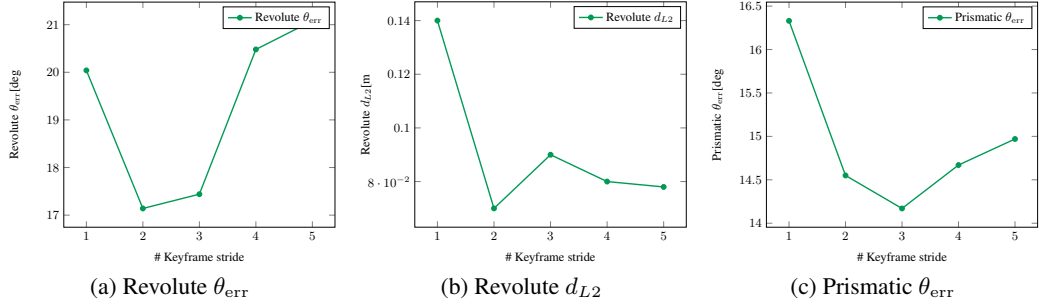
\begin{figure*}
  \centering
  \footnotesize
  \setlength{\tabcolsep}{0.1cm}%
  \begin{tabular}{P{0.32\linewidth}P{0.32\linewidth}P{0.32\linewidth}}
    \resizebox{\linewidth}{!}{
      \begin{tikzpicture}
        \begin{axis}[
          xlabel=\# Keyframe stride,
          ylabel=Revolute $\theta_{\rm err}$[deg],
          label style={font=\Large},
          xtick={1,2,3,4,5},
          xticklabel style={font=\Large},
          yticklabel style={font=\Large},
          xtick align=outside,
          ymin=25, ymax=30,
          y label style={at={(axis description cs:0.13,.5)}, anchor=south},
          x label style={at={(axis description cs:0.5,-.08)}, anchor=south},
          grid=both,
          grid style={line width=.1pt, draw=gray!60},
          legend style={font=\Large, at={(0.5,1.03)}, anchor=south, legend columns=1}
        ]
          \addplot[
            mark=*,
            line width=1.0pt,
            mark options={solid, scale=0.8},
            color=ForestGreen
          ] coordinates {
            (1, 20.04)
            (2, 17.14)
            (3, 17.44)
            (4, 20.48)
            (5, 21.08)
          };
          \addlegendentry{Revolute $\theta_{\rm err}$}
        \end{axis}
      \end{tikzpicture}
    }
    &
    \resizebox{\linewidth}{!}{
      \begin{tikzpicture}
        \begin{axis}[
          xlabel=\# Keyframe stride,
          ylabel=Revolute $d_{L2}$[m],
          label style={font=\Large},
          xtick={1,2,3,4,5},
          xticklabel style={font=\Large},
          yticklabel style={font=\Large},
          xtick align=outside,
          ymin=0.25, ymax=0.45,
          y label style={at={(axis description cs:0.13,.5)}, anchor=south},
          x label style={at={(axis description cs:0.5,-.08)}, anchor=south},
          grid=both,
          grid style={line width=.1pt, draw=gray!60},
          legend style={font=\Large, at={(0.5,1.03)}, anchor=south, legend columns=1}
        ]
          \addplot[
            mark=*,
            line width=1.0pt,
            mark options={solid, scale=0.8},
            color=ForestGreen
          ] coordinates {
            (1, 0.14)
            (2, 0.07)
            (3, 0.09)
            (4, 0.08)
            (5, 0.078)
          };
          \addlegendentry{Revolute $d_{L2}$}
        \end{axis}
      \end{tikzpicture}
    }
    &
    \resizebox{\linewidth}{!}{
      \begin{tikzpicture}
        \begin{axis}[
          xlabel=\# Keyframe stride,
          ylabel=Prismatic $\theta_{\rm err}$[deg],
          label style={font=\Large},
          xtick={1,2,3,4,5},
          xticklabel style={font=\Large},
          yticklabel style={font=\Large},
          xtick align=outside,
          ymin=17, ymax=20,
          y label style={at={(axis description cs:0.13,.5)}, anchor=south},
          x label style={at={(axis description cs:0.5,-.08)}, anchor=south},
          grid=both,
          grid style={line width=.1pt, draw=gray!60},
          legend style={font=\Large, at={(0.5,1.03)}, anchor=south, legend columns=1}
        ]
          \addplot[
            mark=*,
            line width=1.0pt,
            mark options={solid, scale=0.8},
            color=ForestGreen
          ] coordinates {
            (1, 16.33)
            (2, 14.55)
            (3, 14.17)
            (4, 14.67)
            (5, 14.97)
          };
          \addlegendentry{Prismatic $\theta_{\rm err}$}
        \end{axis}
      \end{tikzpicture}
    }
    \\  %
    (a) Revolute $\theta_{\rm err}$ & (b) Revolute $d_{L2}$ & (c) Prismatic $\theta_{\rm err}$
    \end{tabular}
    \caption{Impact of keyframe stride on tracking accuracy and error. (a) The revolute joint angular error $\theta_{\rm err}$[deg] exhibits a minimum at stride 2. (b) The revolute joint positional error $d_{L2}$[m] is lowest at a stride of 2, with higher errors observed for both smaller and larger strides. (c) Similarly, the prismatic joint angular error $\theta_{\rm err}$ exhibits its minimum at stride of 2. We conclude that a stride of 2 is optimal in terms of point density and computational efficiency.}
  \label{fig:keyframe-stride}
\end{figure*}

In this section, we present ablation results on the any-point tracking component as illustrated in \cref{fig:keyframe-stride}. In particular, we evaluate to which degree different keyframe strides used as input to the point tracking component affect the downstream axis prediction performance (see \cref{subsec:point_tracking}). Choosing the keyframe stride is vital hyperparameter of the point tracking stage. While a smaller stride leads to an increase in the number of detected points to be tracked by Cotracker3~\cite{karaev2024cotracker3}, larger strides reduce the number of detected points to be tracked, thereby lowering the computational load. Retaining a sufficient number of points is necessary for estimating an object's 3D trajectory over time. Overall, we observe the lowest prediction errors for a keyframe stride of 2 to 3 based on the angular errors. However, in the case of the translational errors, we are not able to derive a distinct statement. 

\subsection{Scene-Respective Results}
In addition, to the \texttt{EASY}/\texttt{HARD} differentiation evaluated in \cref{tab:difficulty-results}, we show scene-respective results in \cref{tab:scene-results}. This involves averaging the predictions of all object interactions contained in a scene split. We list the number of sequences per scene as well as the number of labeled objects per sequence in \cref{tab:arti4d-sequence-overview}. As reported in \cref{tab:scene-results}, \texttt{RH078} constitutes the most difficult split of the Arti4D dataset. In comparison, \texttt{DR080} and \texttt{RH201} seem to represent simpler environments.

We attribute worse results on \texttt{RH078} to a number of non-separable interactions as the hand is occasionally not fully retrieved between interactions. As a consequence, the proposed interaction extraction baseline potentially fails at differentiating two different interactions. We have mentioned the hand trigger limitation in \cref{sec:Limitations} and leave improvements on that front to feature work. In addition to that, we observe that there are comparably more revolute joints in scene \texttt{RH078} whose associated objects are rather textureless and of metallic character, thus hindering consistent depth observations.

\begin{table}[h]
    \centering
    \scriptsize
    \setlength{\tabcolsep}{8pt}
    \caption{Scene-respective results: We report the scene-wise results using the established set of metrics. We find that \texttt{RH078} constitutes the most difficult split of the Arti4D dataset.}
    \begin{tabular}{l|cccccc}
    \toprule
      \multirow{2}{*}{Method} & \multicolumn{2}{c}{Prismatic joints} & \multicolumn{2}{c}{Revolute joints} & \multicolumn{2}{c}{Type accuracy} \\
       & $\theta_{err}$[deg] & $d_{L2}$[m] & $\theta_{err}$[deg] & $d_{L2}$[m] & Prismatic & Revolute \\
    \midrule
      \texttt{RH078} &  35.11 & -- & 10.37 & 0.05 & 0.55 & 1.00 \\
      \texttt{RR080} &  15.11 & -- & 9.41 & 0.10 & 0.56 & 1.00 \\
      \texttt{DR080} &  7.86 & -- & 17.97 & 0.12 & 0.83 & 1.00 \\
      \texttt{RH201} &  8.73 & -- & 20.88 & 0.04 & 0.84 & 0.95 \\
      \midrule %
      Overall &  \textbf{14.55} & -- & \textbf{17.14} & \textbf{0.07} & \textbf{0.68} & \textbf{0.98} \\
    \bottomrule
    \end{tabular}
    \label{tab:scene-results}
\end{table}

\subsection{Qualitative Results}
\label{sec:qualitative-results-suppl}
\begin{figure}
    \centering
    \begin{tabular}{cc}
        \includegraphics[width=0.5\linewidth, clip, trim={5cm 0 0 0}]{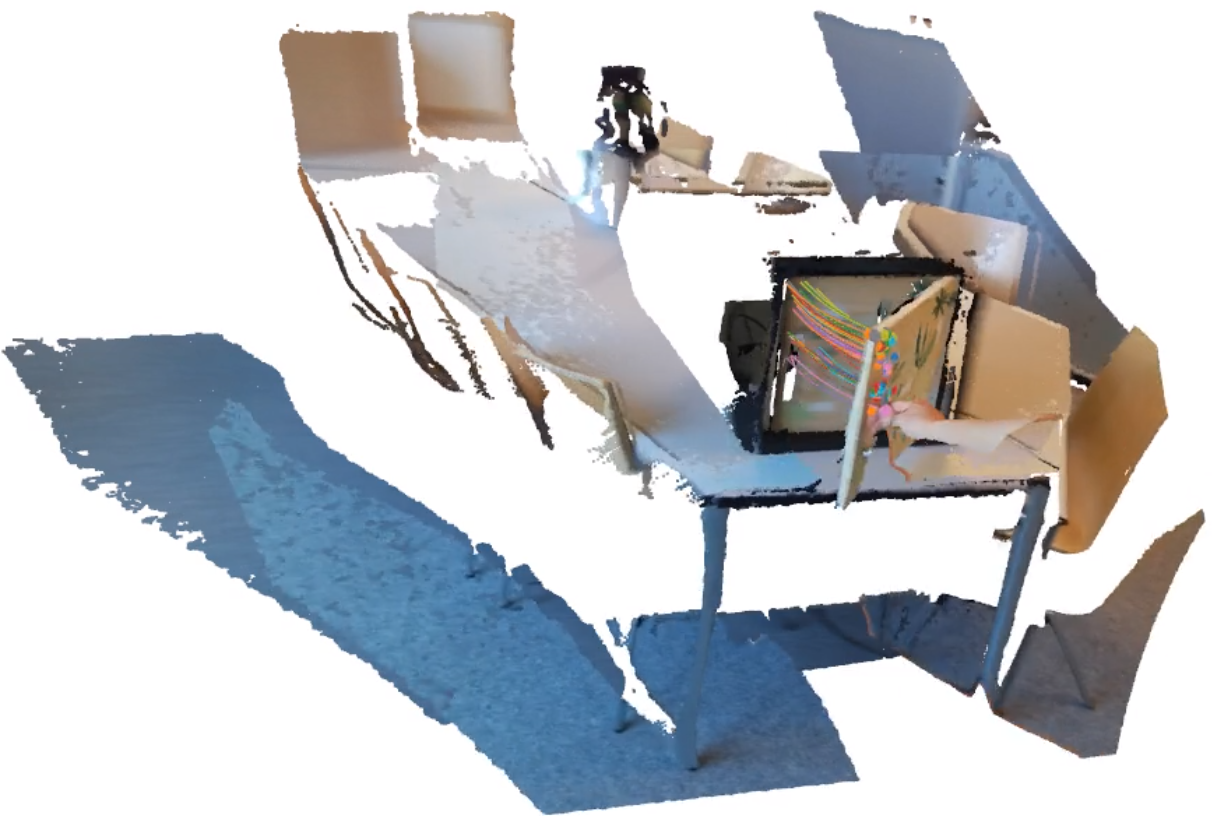}  & \includegraphics[width=0.45\linewidth, clip, trim={0 0 8cm 0}]{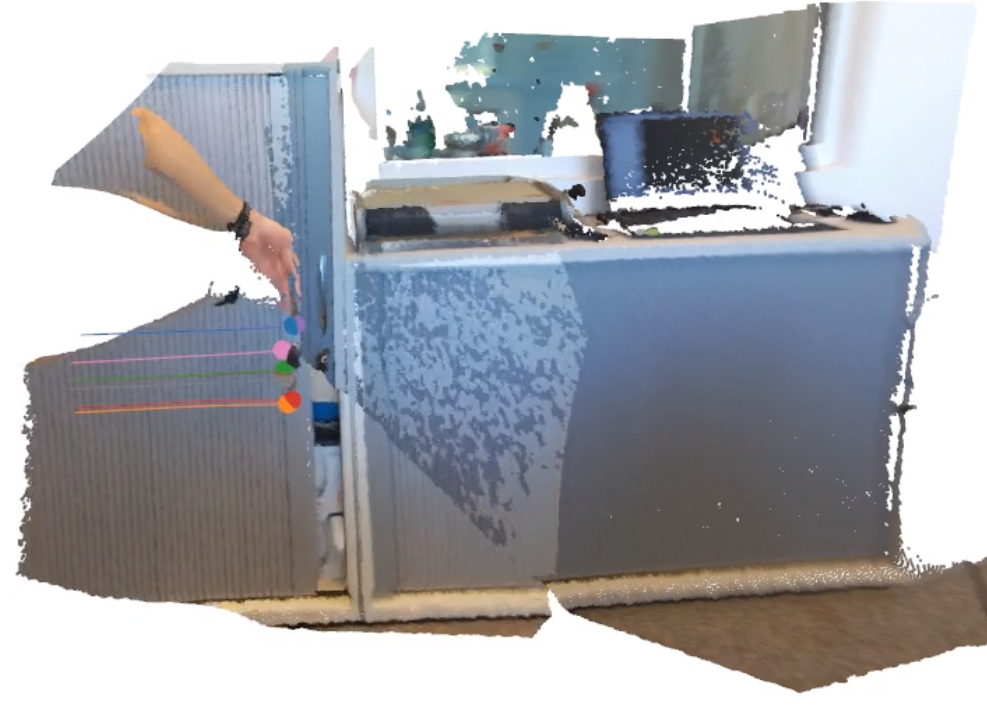} \\
    \end{tabular}
    
    \caption{Smoothed point trajectories: We visualize a cabinet (revolute) and its tracked keypoints (left) as well as a linear slider shelf (prismatic) on the right. Each keypoint trajectory is represented with a unique color. Both sets of point trajectories visualized constitute the output of our track filtering introduced in \cref{subsec:3d_track_estimation}.}
    \label{fig:point-traj}
\end{figure}

\begin{figure}
    \centering
    \includegraphics[width=1.0\linewidth, clip, trim={0 0 0 5cm}]{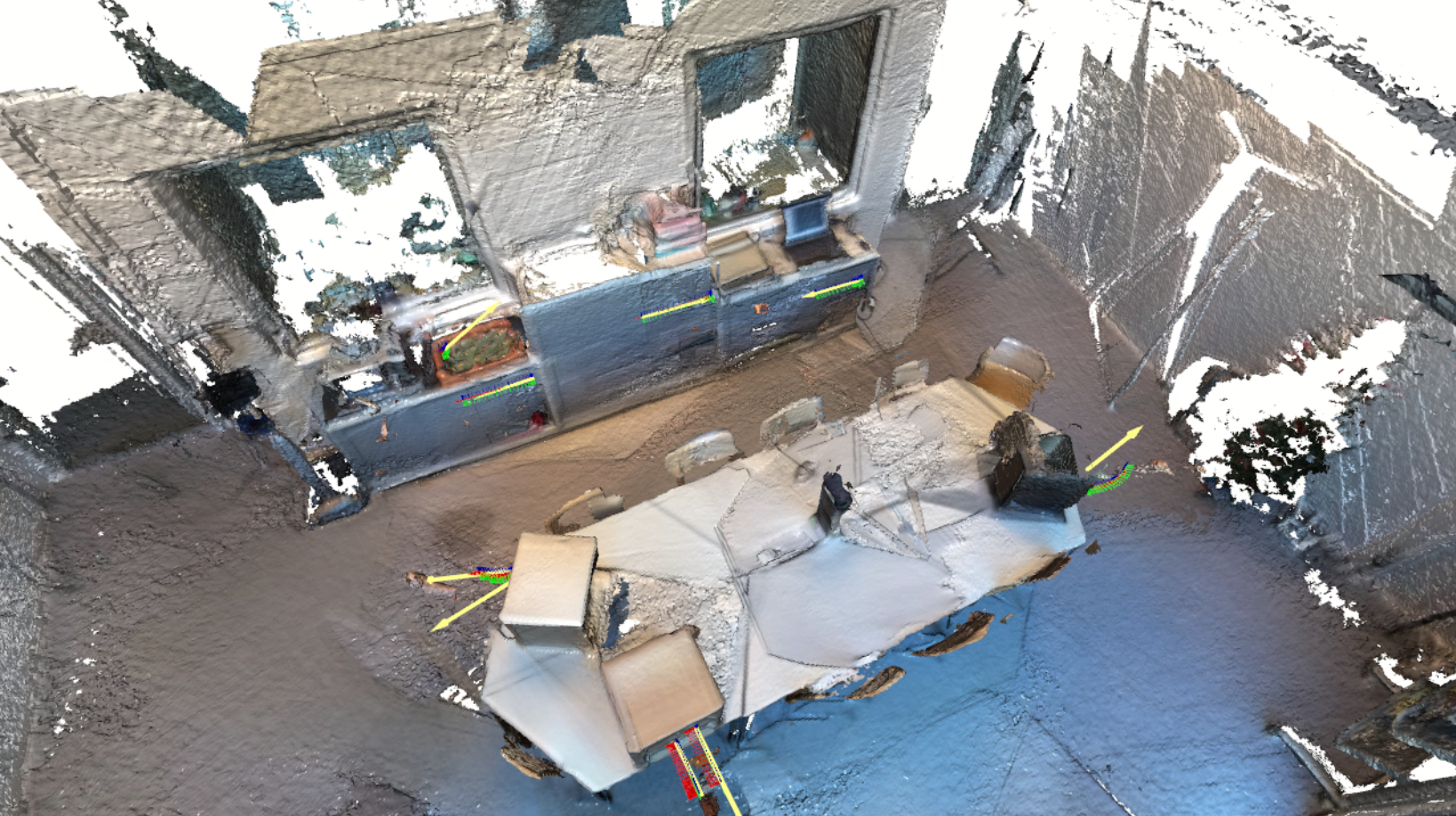}
    \caption{Scene-level prediction: We depict a full scene-level output of the ArtiPoint framework on sequence \texttt{scene\_2025-04-11-11-44-32} of the \texttt{DR080} scene. Yellow arrows denote axes of motion of predicted object interactions while coordinate frames represent the estimate part poses throughout articulation based on the proposed estimation framework (see \cref{subsec:articulation_estimation}).}
    \label{fig:scene-level}
\end{figure}

In the following, we provide additional qualitative results. In \cref{fig:point-traj}, we visualize the output of our proposed interaction extraction, any-point tracking and track filtering components. We observe a sufficient number of point trajectories even under partially missing depth measurements or feature-sparse textures. In addition to that, we visualize a full scene-level output of ArtiPoint on \texttt{DR080} scene in \cref{fig:scene-level}. Overall, ArtiPoint detects the majority of interactions and produces reliable estimates considering the in-the-wild character of the recorded Arti4D sequences.

\section{Arti4D Dataset}
\label{sec:arti4d-suppl}
In the following, we provide additional insight on the in-the-wild object articulation dataset Arti4D. We provide 45 sequences across four distinct environments as listed in \cref{tab:arti4d-sequence-overview} and visualized in \cref{fig:arti4d-examples}. In addition to the sequence IDs and the recording names of the produced sequences, we report the number of labeled objects for each sequence, the ratio between prismatic and revolute joints as well as the ratio between easy and hard objects. First, note that the number of objects is not equal to the number of object interactions per sequence, as several sequences contain repeated interactions with the same single object instance. This constitutes a corner case in terms of articulated object interaction as it requires prediction methods to fuse, e.g., two predictions belonging to a single object. While most interactions are separable by detecting the absence of a hand mask, especially the \texttt{RH078} split which contains a number of hard-to-separate interactions. This is due to the fact that the interacting hand was not always fully retrieved in-between interactions of two distinct objects. Similar to the repeated interactions case mentioned before, this represents another corner case requiring advanced action recognition.

\begin{figure*}
\centering
\setlength{\tabcolsep}{0.1cm}%
{
\renewcommand{\arraystretch}{0.5}%

\begin{tabular}{cc}

{\includegraphics[width=0.5\linewidth]{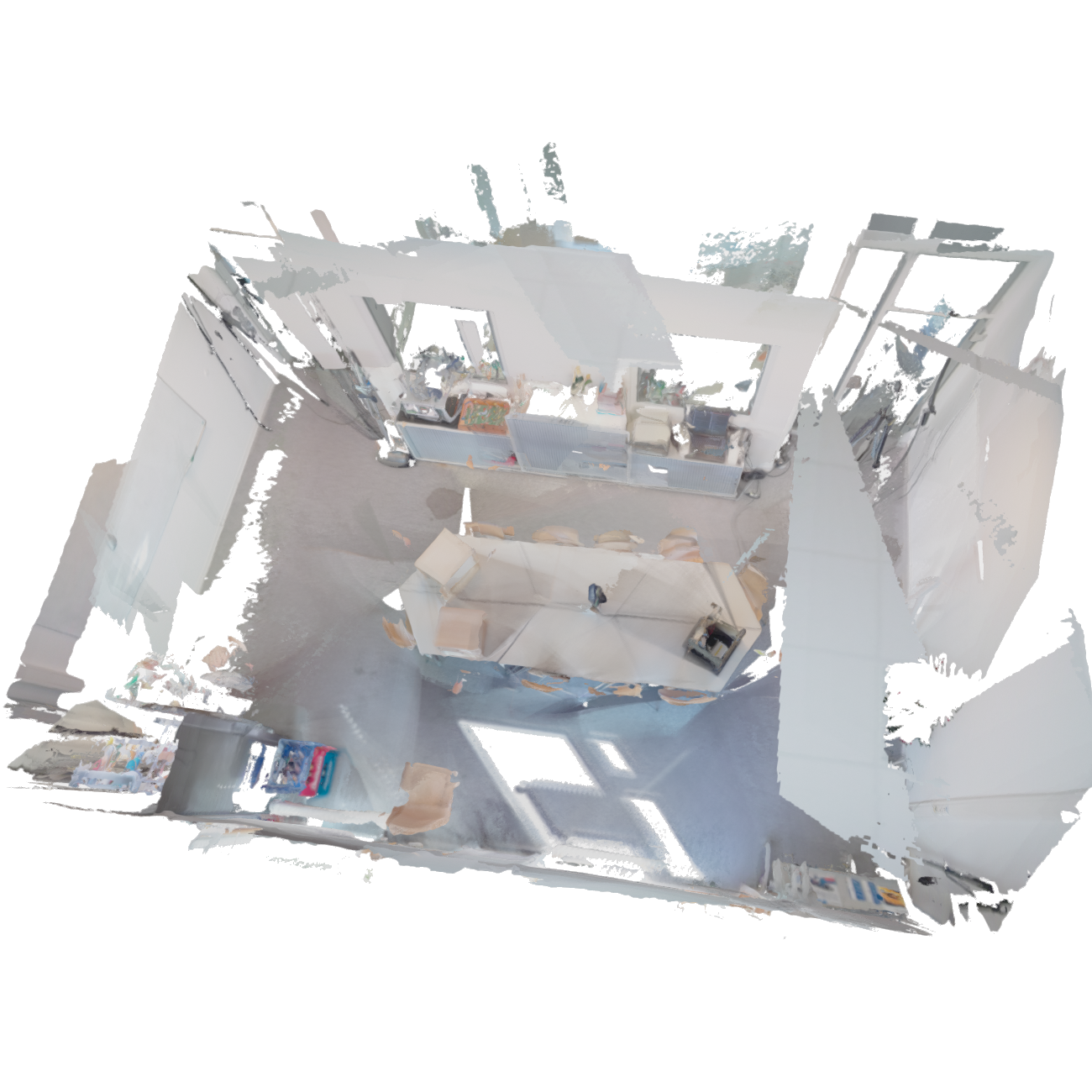}} & 
{\includegraphics[width=0.5\linewidth]{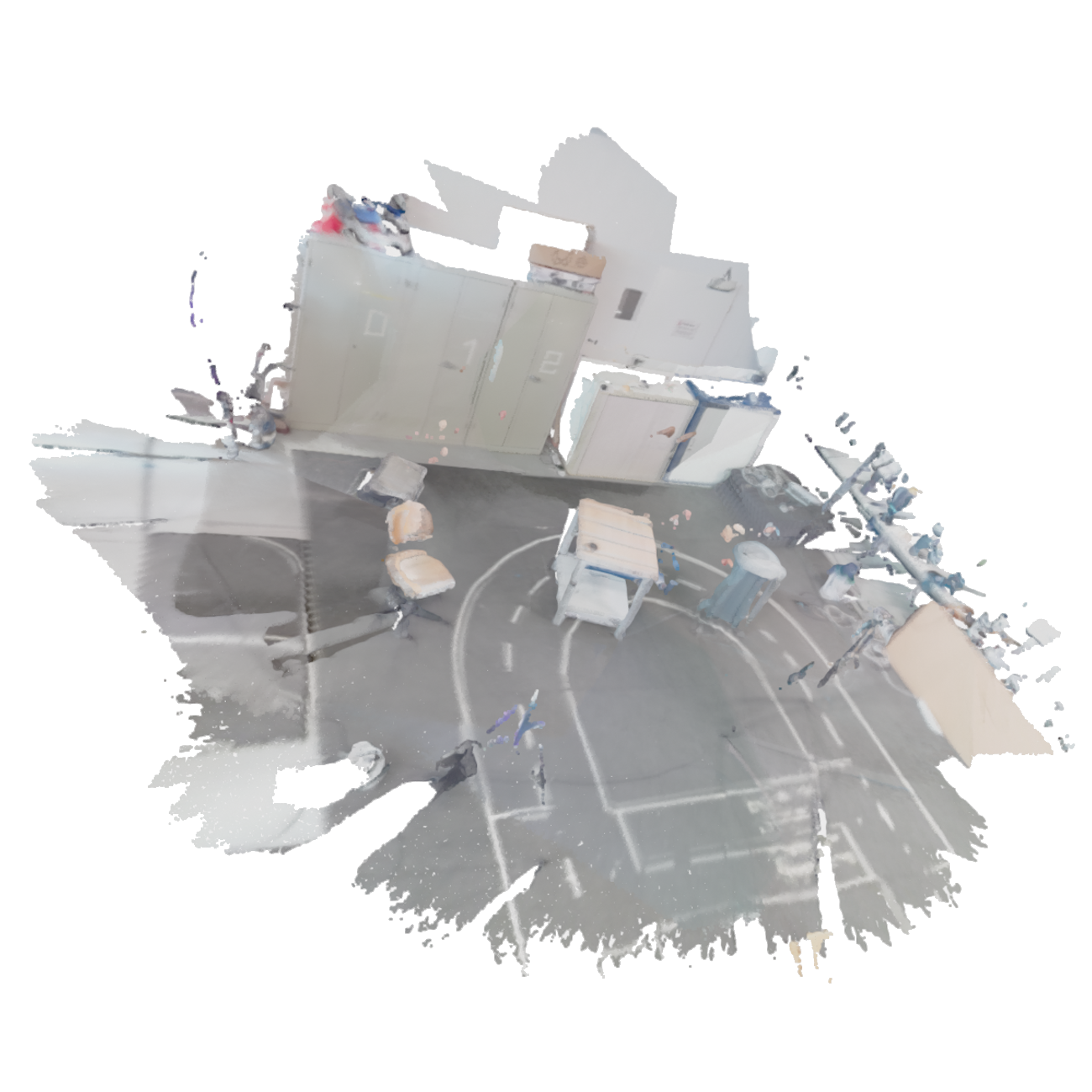}} \\
\texttt{DR080} & \texttt{RH078} \\

{\includegraphics[width=0.5\linewidth]{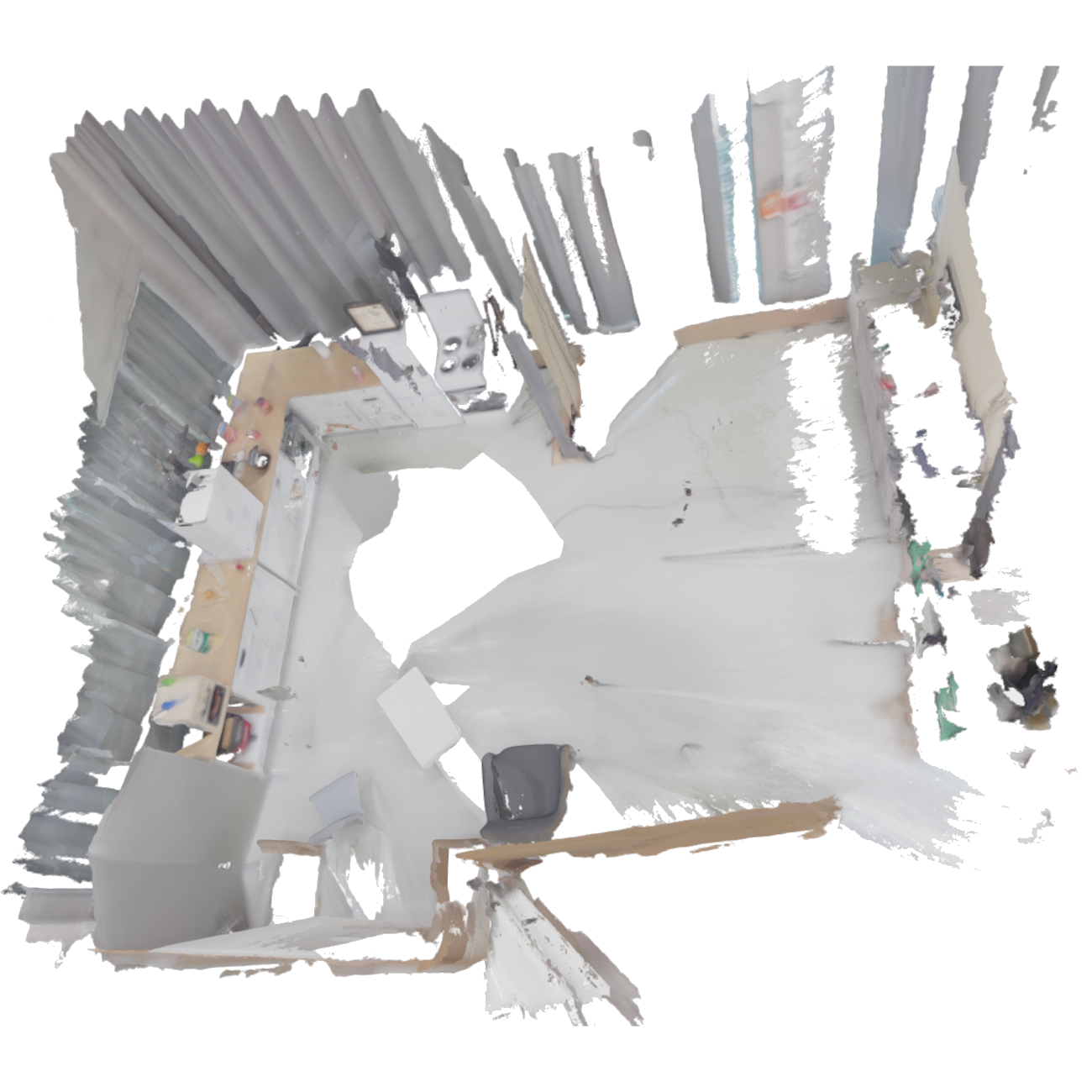}} & 
{\includegraphics[width=0.5\linewidth]{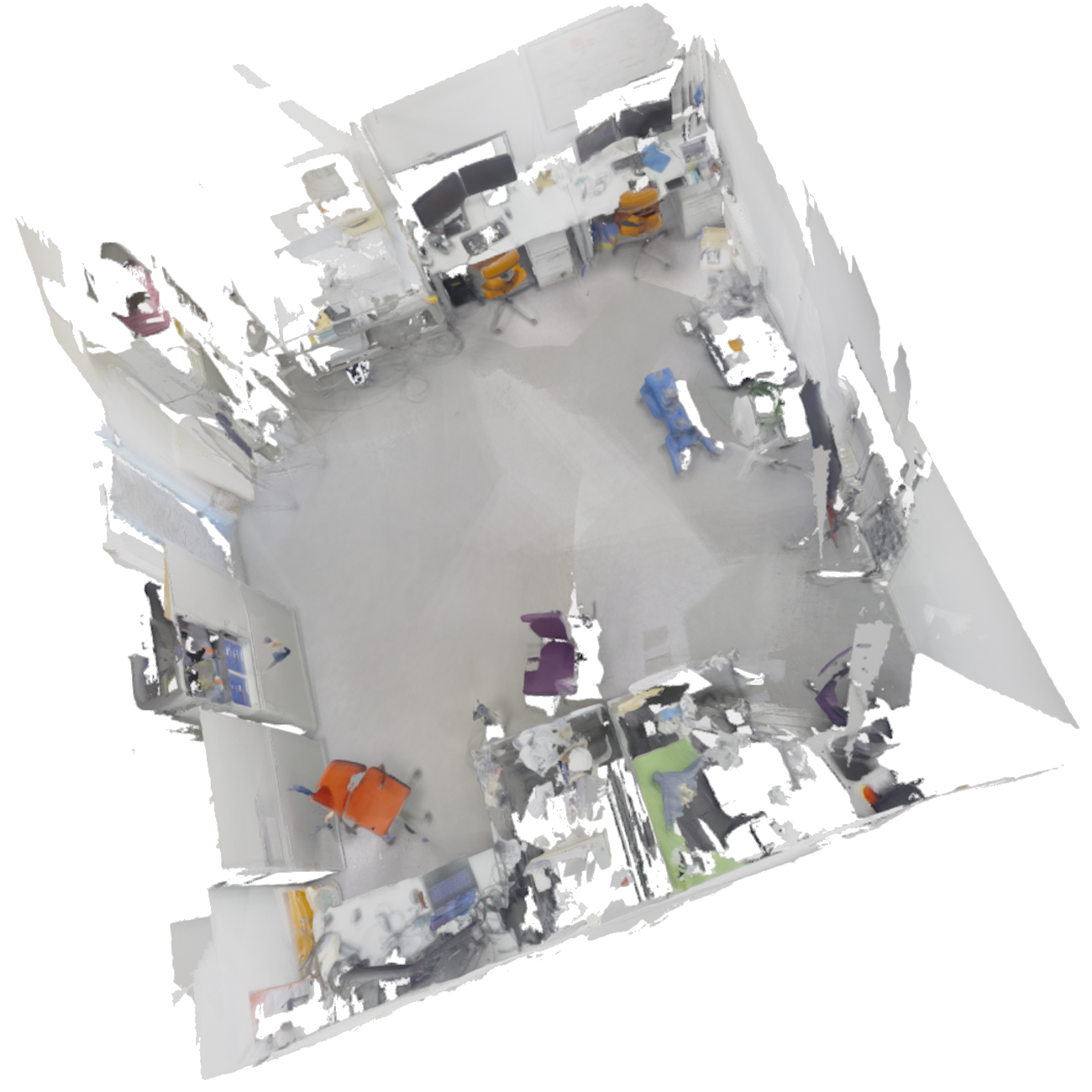}} \\
\texttt{RH201} & \texttt{RR080} \\

\end{tabular}
}
\caption{We visualize the reconstructed scenes of the four Arti4D environments: \texttt{DR080}, \texttt{RH078}, \texttt{RH201}, and \texttt{RR080}.}
\label{fig:reconstructed-seqs}

\end{figure*}

As part of the dataset, we make both rosbags and processed raw data public. While the rosbags include TF data at a higher frequency, the raw data includes aligned RGB, depth, and camera poses at 15~Hz. We employed an Azure Kinect RGB-D camera that was handheld throughout all interactions. In terms of ground truth camera pose retrieval, we employed external tracking using HTC Vive trackers, which provide cm-level accuracy. In case of sudden odometry glitches induced by considerable occlusions or reflections on glass or metal, we have removed those sequences from the dataset. We found that running classical structure-from-motion approaches to reconstruct the underlying sequence fails as significant parts of the camera field of view cover articulations. In turn, the contained articulations break assumptions towards mostly \textit{static} visual correspondence made in structure-from-motion methods. Thus, we leverage the ground truth camera poses and perform TSDF fusion to produce scene reconstructions. We depict four TSDF-reconstructed sequences stemming from each of the splits in \cref{fig:reconstructed-seqs}. The reconstructions reflect minimal ground truth odometry drift and enable precise anchoring of object axes. 

\begin{figure*}[h!]
\centering
\footnotesize
\setlength{\tabcolsep}{0.03cm}%

\begin{tabularx}{\linewidth}{ccc}
\includegraphics[width=0.5\linewidth]{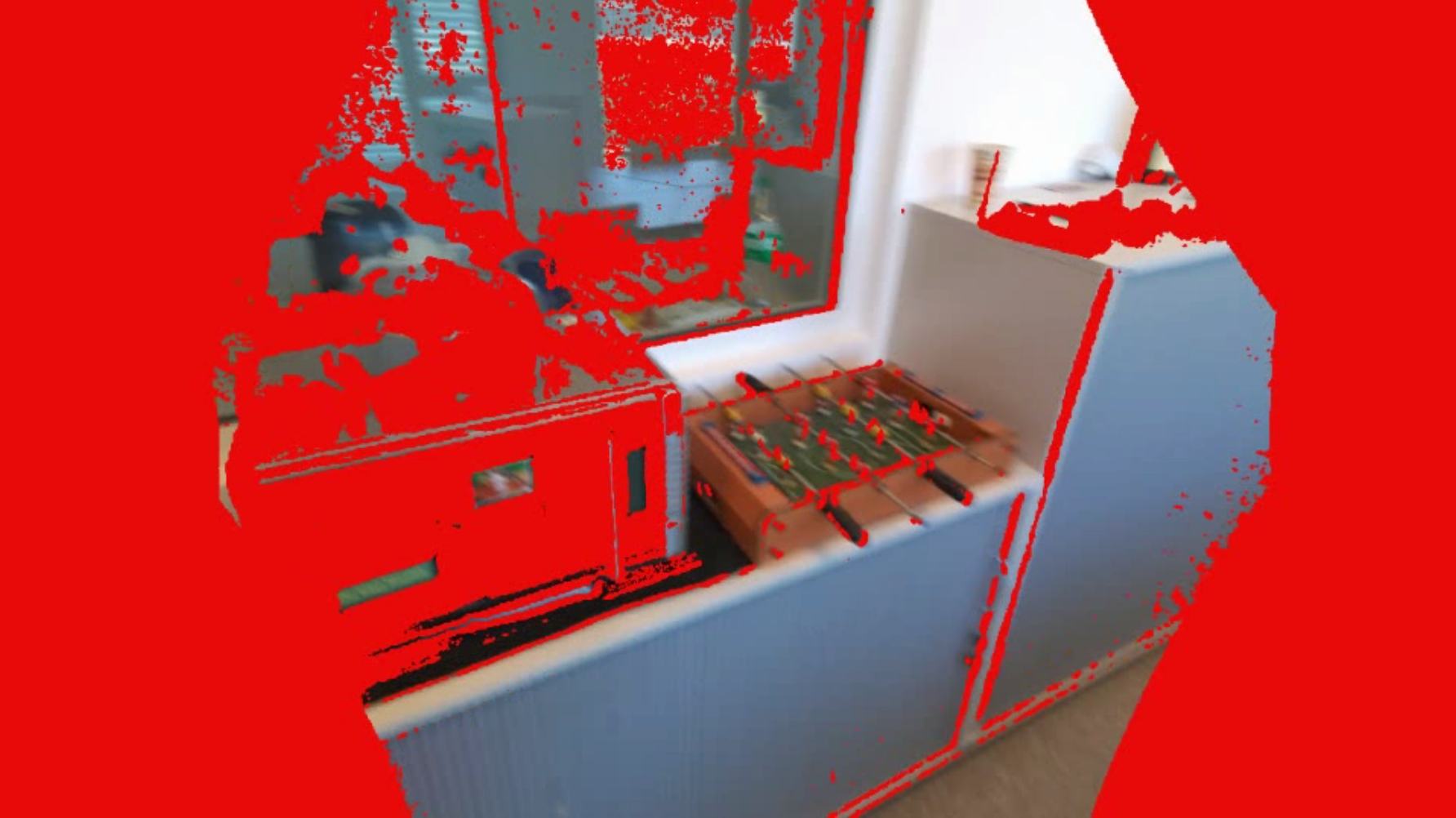}  & \includegraphics[width=0.5\linewidth]{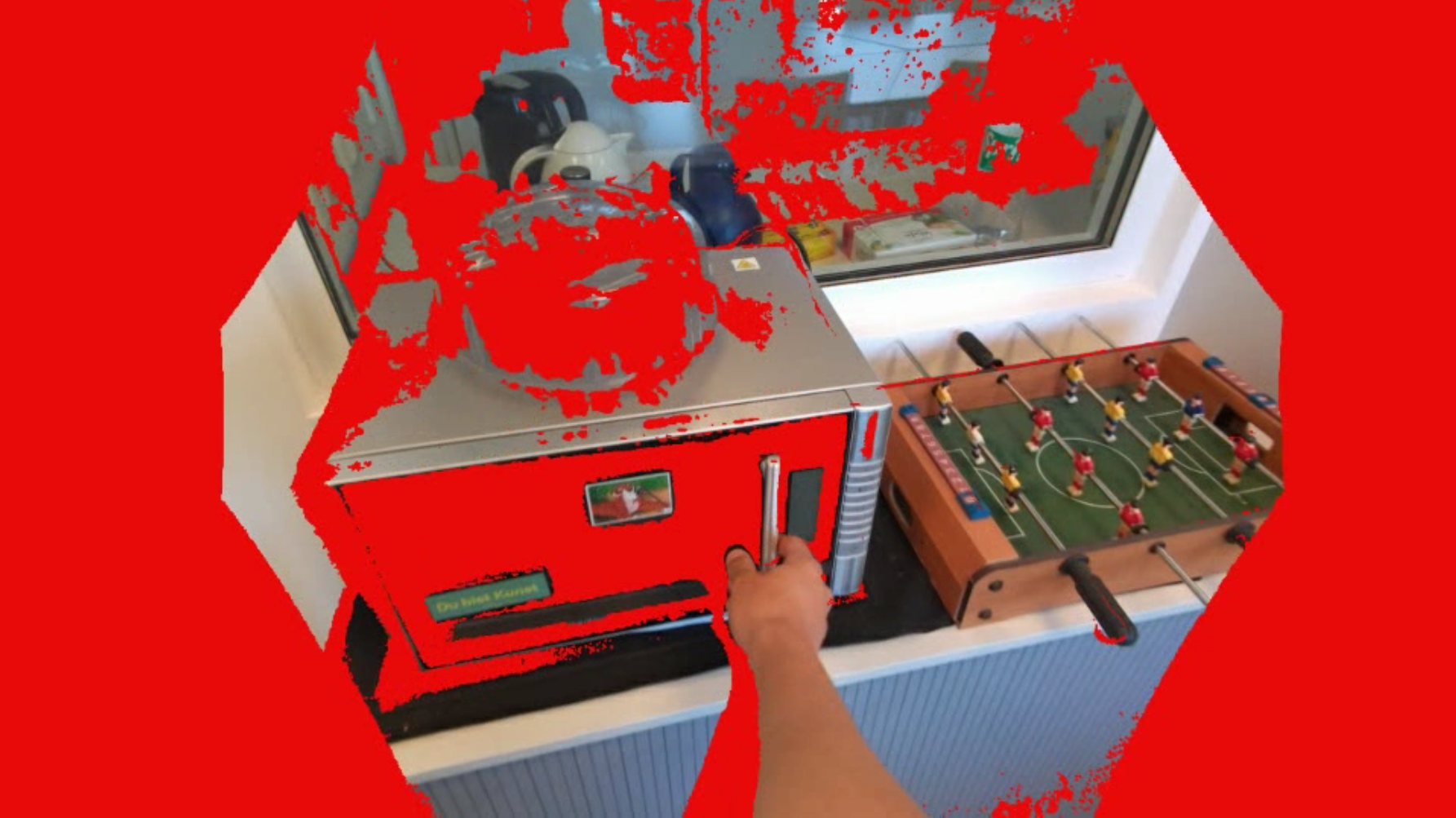} \\[-1pt]
\includegraphics[width=0.5\linewidth]{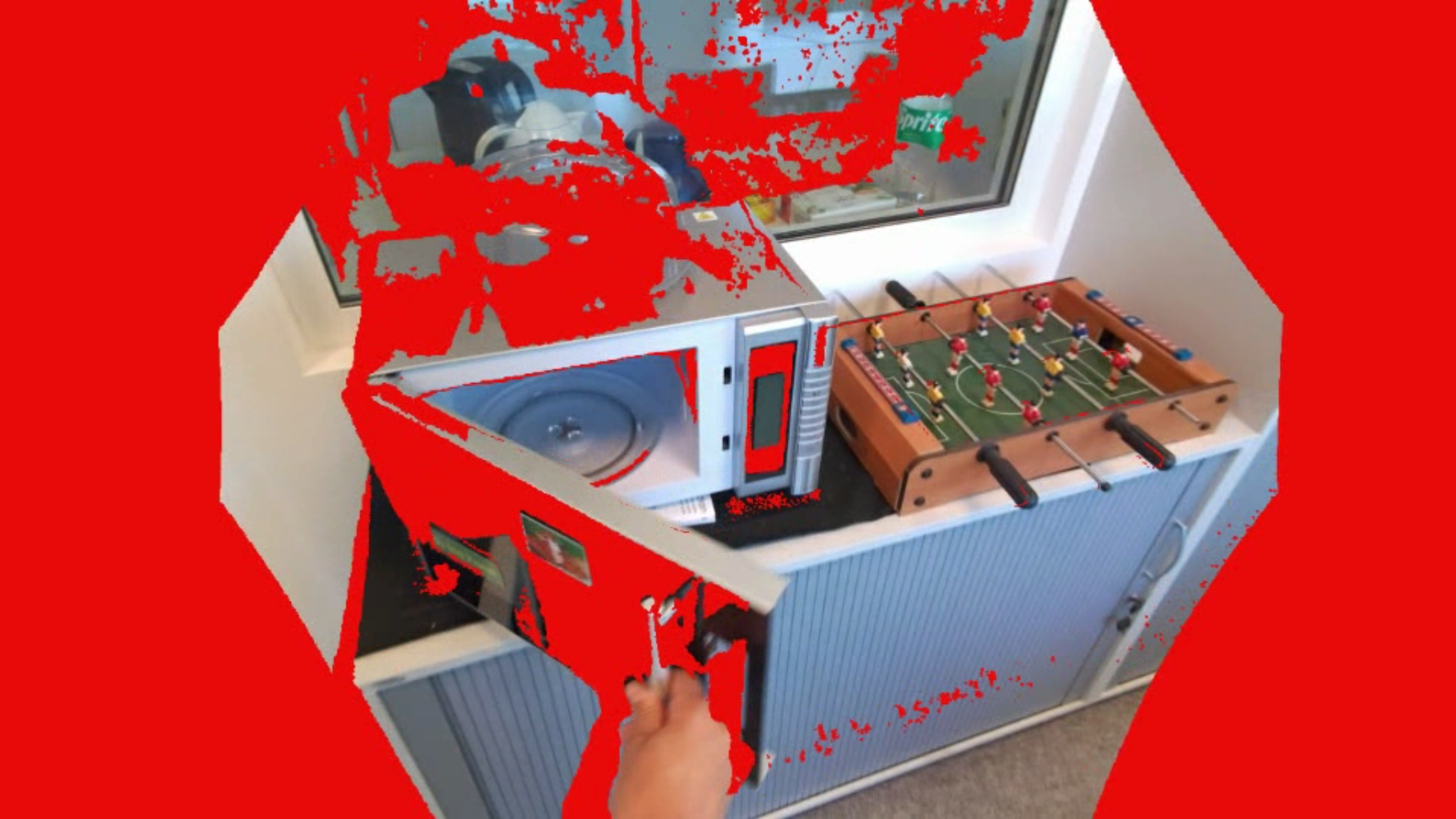} &
\includegraphics[width=0.5\linewidth]{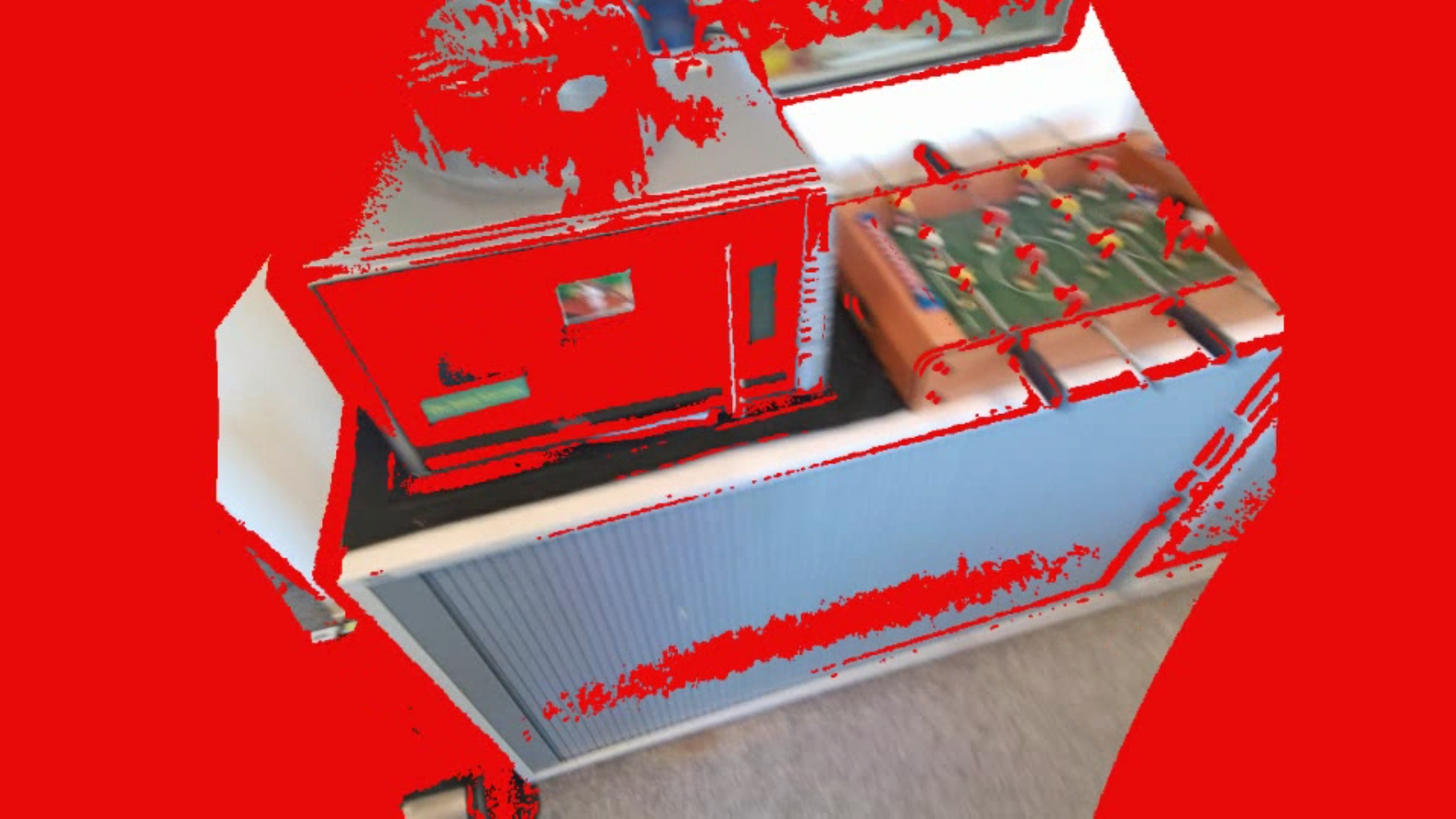} \\
\end{tabularx}
\caption{We visualize an exemplary interaction with the depth projected onto the RGB observation in red wherever a depth reading was unavailable for the particular image coordinate. As depicted, metallic, semi-transparent, or dark-colored objects do not produce reliable depth estimates, ultimately complicating the lifting of 3D point trajectories used to estimate the underlying articulation.}
\label{fig:missing-depth}
\end{figure*}

The ground truth object axes were labeled based on the reconstructed sequences using Blender, exported as JSONs, and verified by a second reviewer. Furthermore, we provide metadata on difficulty levels and temporal interaction segment ground truth labels. We provide four exemplary depth-masked RGB frames covering an interaction of a microwave featuring semi-transparent glass and metal surfaces in \cref{fig:missing-depth}. As depicted, a considerable part of the object does not produce depth estimates. Objects of that kind are labeled using the \texttt{HARD} category.

Finally, we depict several RGB frames drawn from the four distinct scene splits in \cref{fig:arti4d-examples}, underlining the variety of objects and challenging conditions of the proposed dataset. We also make the sequence reconstructions in the form of meshes and point clouds available.

\begin{table*}[ht]
    \footnotesize
    \setlength{\tabcolsep}{6pt}
    \centering
    \label{tab:arti4d-sequence-overview}
    \caption{Overview of all Arti4D demonstration sequences: We assign sequence identifiers to each recording and list the number of objects interacted with per sequence as well as the distribution of prismatic vs. revolute joints (\texttt{PRISM} / \texttt{REV}) and objects classified as either \texttt{EASY} or \texttt{HARD}. The number of objects corresponds to the number of annotated axes per sequence, whereas the number of interactions denotes the number of performed articulations, thus including objects that are articulated multiple times per sequence.}
    \begin{threeparttable}
        \begin{tabularx}{0.99\linewidth}{cclcccc}
          \toprule
            \multirow{2}{*}{\textit{Scene}} & \textit{Sequence} & \multirow{2}{*}{\textit{Recording}} & \textit{\#} & \textit{\#} & \#  & \# \\
                           & \textit{ID} &  & \textit{Objects} & \textit{Interactions} & \texttt{PRISM} / \texttt{REV} &  \texttt{EASY} / \texttt{HARD} \\
             \midrule
                \multirow{11}{*}{\rotatebox{90}{\texttt{RH078}}} & RH078-00 & scene\_2025-04-04-19-14-38 &  7  &  7  &  3 / 4  &  7 / 0  \\
                 & RH078-01 & scene\_2025-04-04-19-18-54 &  6  &  7  &  2 / 4  &  4 / 2  \\
                 & RH078-02 & scene\_2025-04-07-11-39-17 &  7  &  7  &  3 / 4  &  7 / 0  \\
                 & RH078-03 & scene\_2025-04-07-11-41-52 &  8  &  9  &  3 / 5  &  6 / 2  \\
                 & RH078-04 & scene\_2025-04-07-11-48-40 &  7  &  7  &  3 / 4  &  7 / 0  \\
                 & RH078-05 & scene\_2025-04-09-10-30-11 &  8  &  8  &  5 / 3  &  3 / 5  \\
                 & RH078-06 & scene\_2025-04-09-10-32-52 &  6  &  6  &  5 / 1  &  3 / 3  \\
                 & RH078-07 & scene\_2025-04-09-10-35-47 &  7  &  7  &  6 / 1  &  2 / 5  \\
                 & RH078-08 & scene\_2025-04-09-10-38-38 &  7  &  7  &  5 / 2  &  3 / 4  \\
                 & RH078-09 & scene\_2025-04-09-10-46-48 &  7  &  8  &  7 / 0  &  4 / 3  \\
                 & RH078-10 & scene\_2025-04-09-10-49-20 &  7  &  7  &  6 / 1  &  2 / 5  \\
        
                \greyrule
                \multirow{10}{*}{\rotatebox{90}{\texttt{RR080}}} & RR080-00 & scene\_2025-04-10-13-11-16 &  17  &  17  &  14 / 3  &  9 / 8  \\
                 & RR080-01 & scene\_2025-04-10-16-05-09 &  14  &  14  &  13 / 1  &  8 / 6  \\
                 & RR080-02 & scene\_2025-04-17-15-25-14 &  11  &  12  &  11 / 0  &  7 / 4  \\
                 & RR080-03 & scene\_2025-04-17-15-33-44 &  9  &  9  &  9 / 0  &  7 / 2  \\
                 & RR080-04 & scene\_2025-04-22-09-53-49 &  8  &  8  &  8 / 0  &  5 / 3  \\
                 & RR080-05 & scene\_2025-04-22-09-56-24 &  10  &  10  &  9 / 1  &  9 / 1  \\
                 & RR080-06 & scene\_2025-04-22-09-58-49 &  7  &  8  &  7 / 0  &  4 / 3  \\
                 & RR080-07 & scene\_2025-04-22-11-45-15 &  9  &  9  &  8 / 1  &  7 / 2  \\
                 & RR080-08 & scene\_2025-04-22-11-48-01 &  9  &  9  &  8 / 1  &  7 / 2  \\
                 & RR080-09 & scene\_2025-04-22-11-50-40 &  8  &  8  &  7 / 1  &  6 / 2  \\

                 \greyrule
                \multirow{8}{*}{\rotatebox{90}{\texttt{DR080}}} & DR080-00 & scene\_2025-04-11-11-44-32 &  11  &  11  &  7 / 4  &  5 / 6  \\
                 & DR080-01 & scene\_2025-04-11-12-58-58 &  10  &  10  &  6 / 4  &  5 / 5  \\
                 & DR080-02 & scene\_2025-04-11-13-01-59 &  9  &  9  &  5 / 4  &  4 / 5  \\
                 & DR080-03 & scene\_2025-04-11-13-18-00 &  9  &  10  &  4 / 5  &  3 / 6  \\
                 & DR080-04 & scene\_2025-04-11-13-43-03 &  11  &  11  &  7 / 4  &  5 / 6  \\
                 & DR080-05 & scene\_2025-04-11-14-01-06 &  11  &  11  &  7 / 4  &  5 / 6  \\
                 & DR080-06 & scene\_2025-04-11-15-43-24 &  11  &  12  &  7 / 4  &  5 / 6  \\
                 & DR080-07 & scene\_2025-04-11-15-46-48 &  11  &  11  &  6 / 5  &  4 / 7  \\

                \greyrule
                \multirow{16}{*}{\rotatebox{90}{\texttt{RH201}}} & RH201-00 & scene\_2025-04-24-17-52-21 &  11  &  11  &  5 / 6  &  6 / 5  \\
                 & RH201-01 & scene\_2025-04-24-17-54-13 &  9  &  9  &  5 / 4  &  5 / 4  \\
                 & RH201-02 & scene\_2025-04-24-19-18-42 &  11  &  11  &  5 / 6  &  7 / 4  \\
                 & RH201-03 & scene\_2025-04-24-19-21-50 &  8  &  8  &  2 / 6  &  5 / 3  \\
                 & RH201-04 & scene\_2025-04-24-19-24-09 &  8  &  8  &  5 / 3  &  5 / 3  \\
                 & RH201-05 & scene\_2025-04-25-10-36-37 &  9  &  9  &  5 / 4  &  7 / 2  \\
                 & RH201-06 & scene\_2025-04-25-10-53-40 &  8  &  8  &  4 / 4  &  3 / 5  \\
                 & RH201-07 & scene\_2025-04-25-10-56-33 &  8  &  8  &  3 / 5  &  3 / 5  \\
                 & RH201-08 & scene\_2025-04-25-11-11-47 &  15  &  16  &  6 / 9  &  6 / 9  \\
                 & RH201-09 & scene\_2025-04-25-11-15-47 &  7  &  7  &  5 / 2  &  4 / 3  \\
                 & RH201-10 & scene\_2025-04-25-14-58-42 &  9  &  9  &  6 / 3  &  4 / 5  \\
                 & RH201-11 & scene\_2025-04-25-15-02-14 &  7  &  7  &  3 / 4  &  4 / 3  \\
                 & RH201-12 & scene\_2025-04-25-15-04-48 &  7  &  7  &  4 / 3  &  3 / 4  \\
                 & RH201-13 & scene\_2025-04-25-15-16-29 &  9  &  9  &  5 / 4  &  3 / 6  \\
                 & RH201-14 & scene\_2025-04-25-15-19-22 &  10  &  10  &  5 / 5  &  5 / 5  \\
                 & RH201-15 & scene\_2025-04-25-15-22-54 &  8  &  8  &  4 / 4  &  3 / 5  \\         
            \bottomrule
        \end{tabularx}
    \end{threeparttable}
\end{table*}

\begin{figure*}
\centering
\setlength{\tabcolsep}{0.1cm}%
{
\renewcommand{\arraystretch}{0.5}%

\begin{tabular}{llll}
\multirow{2}{*}{\texttt{DR080}} & 
{\includegraphics[width=0.3\linewidth]{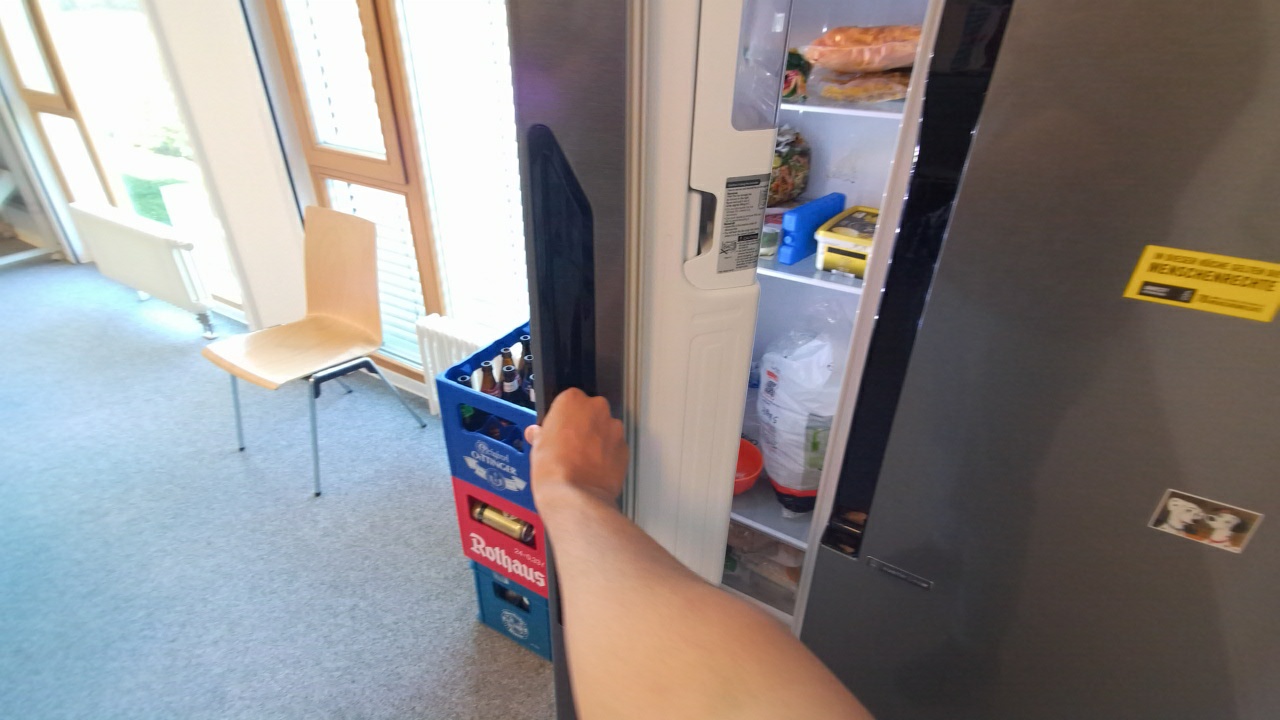}} & 
{\includegraphics[width=0.3\linewidth]{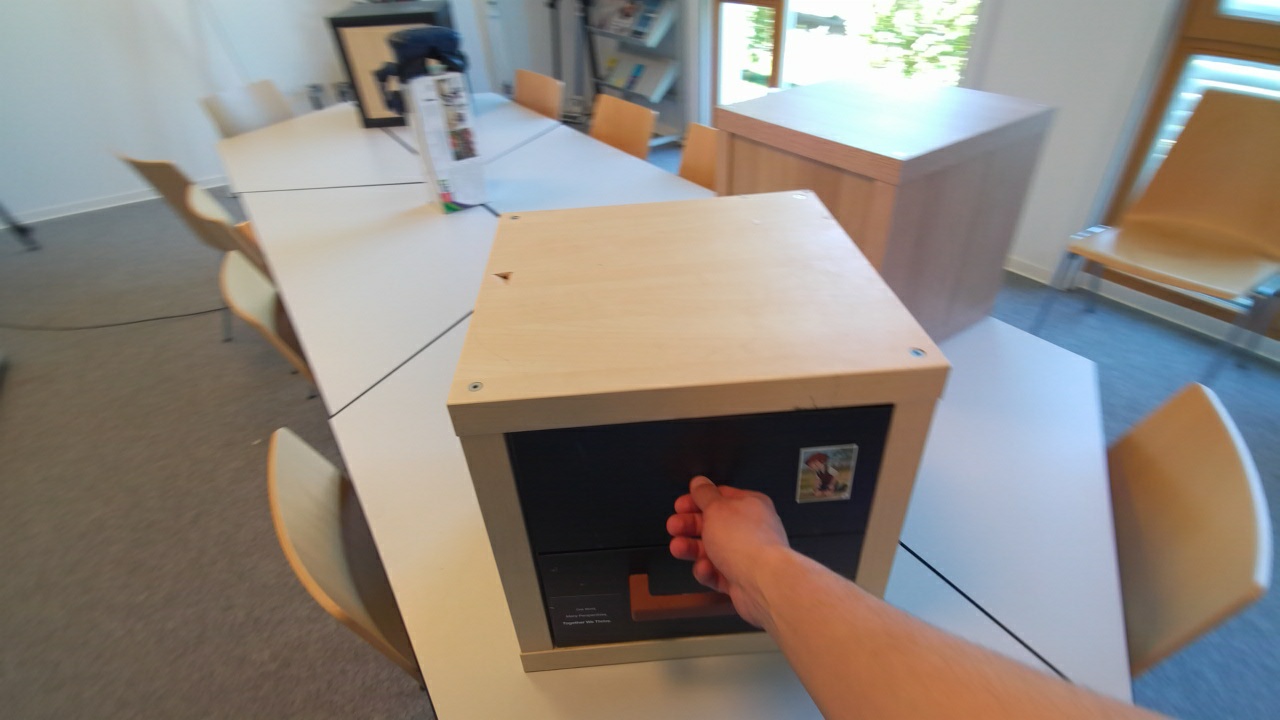}} & 
{\includegraphics[width=0.3\linewidth]{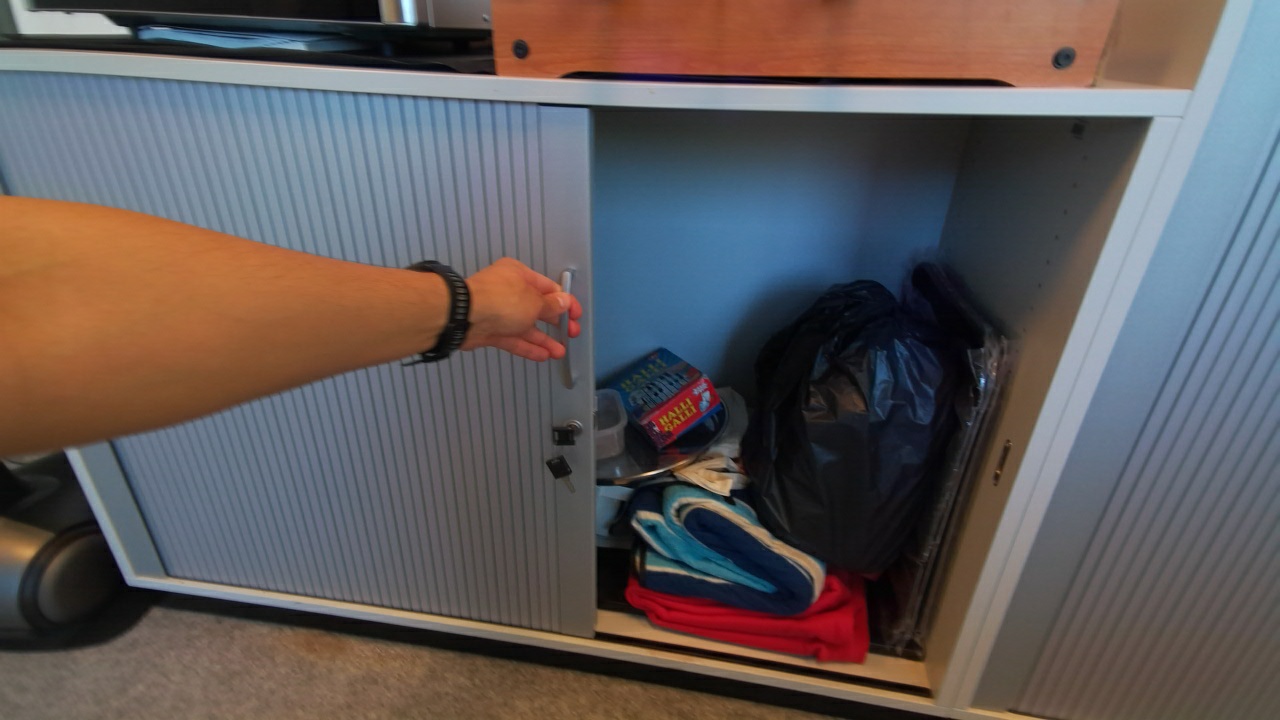}} \\
 & 
{\includegraphics[width=0.3\linewidth]{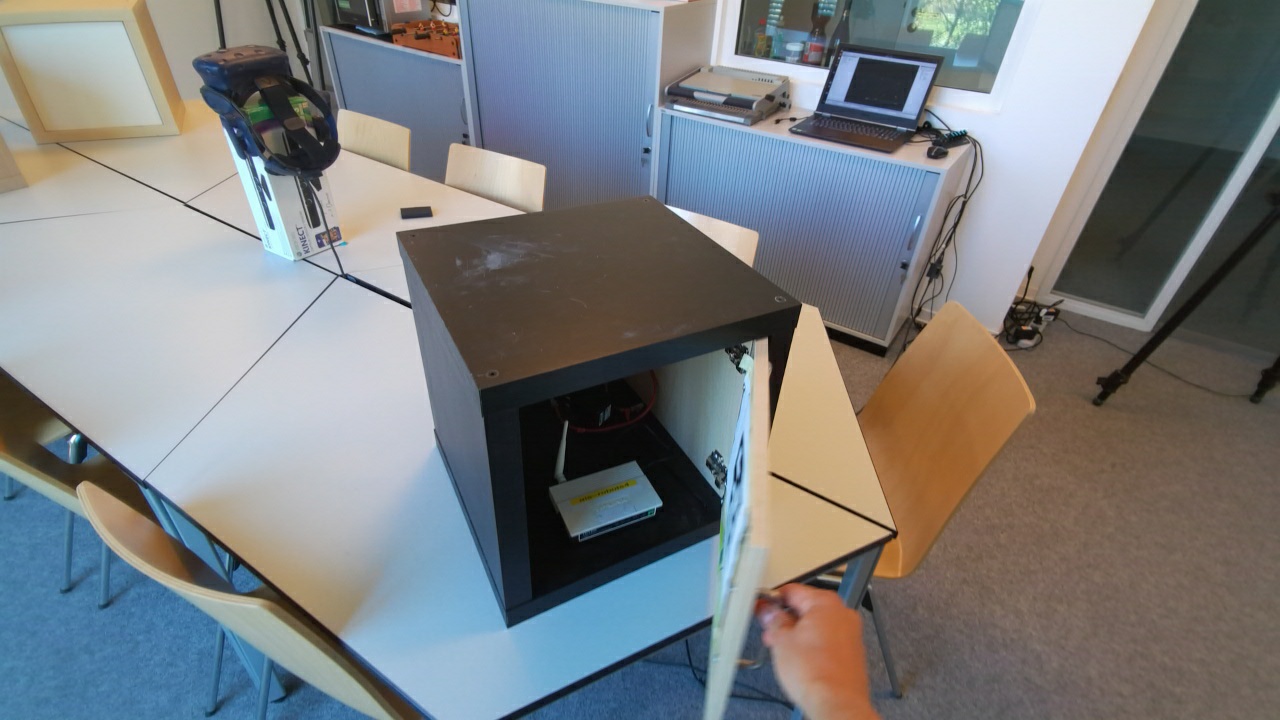}} & 
{\includegraphics[width=0.3\linewidth]{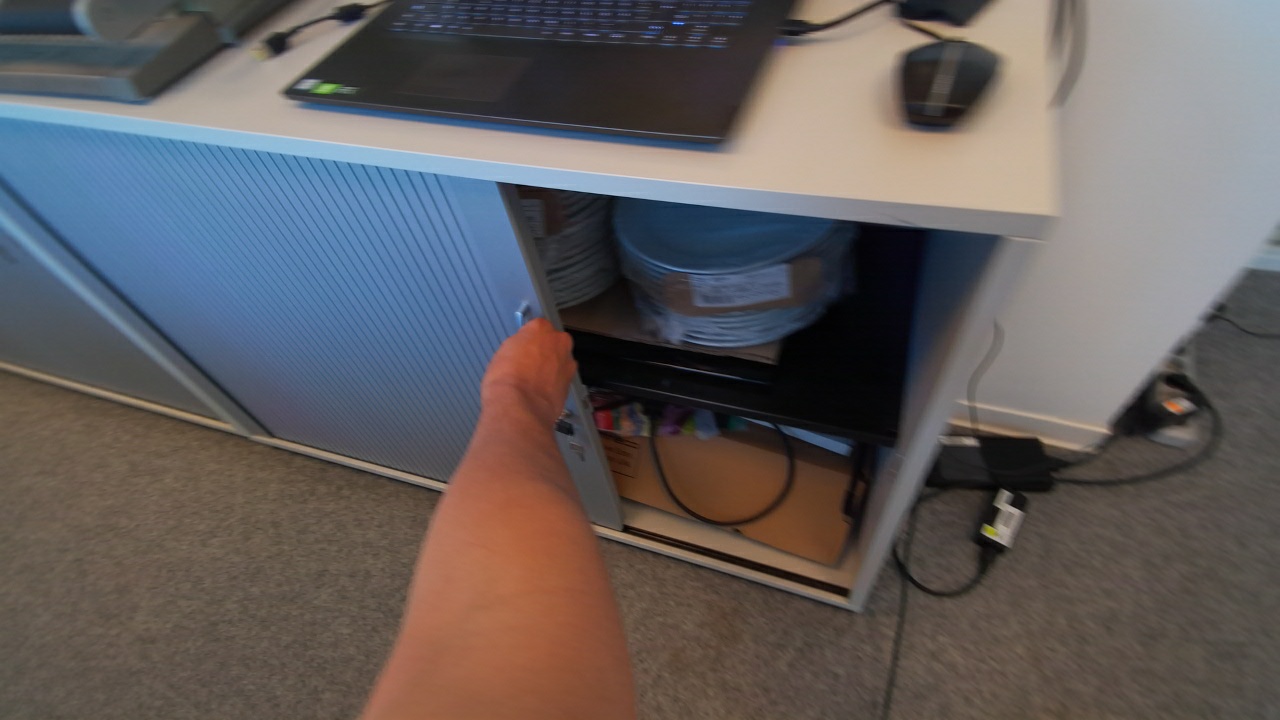}} & 
{\includegraphics[width=0.3\linewidth]{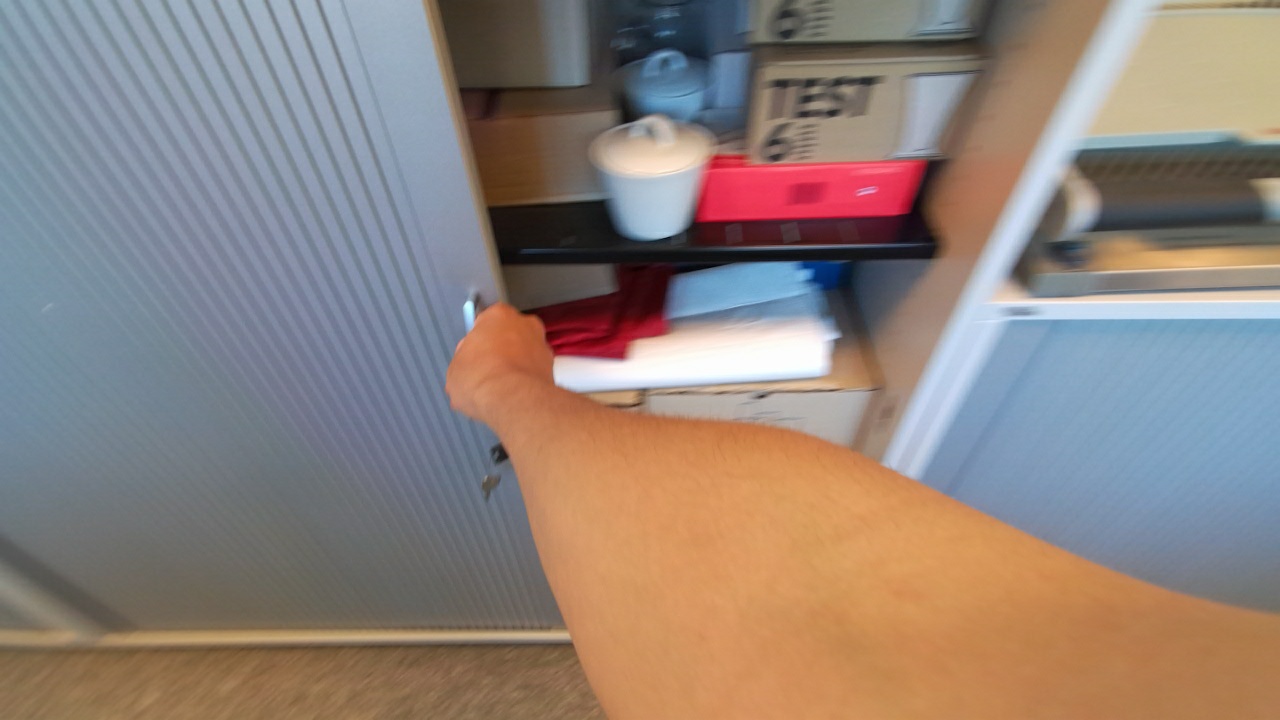}} \\
\\
\\

\multirow{2}{*}{\texttt{RH078}} & 
{\includegraphics[width=0.3\linewidth]{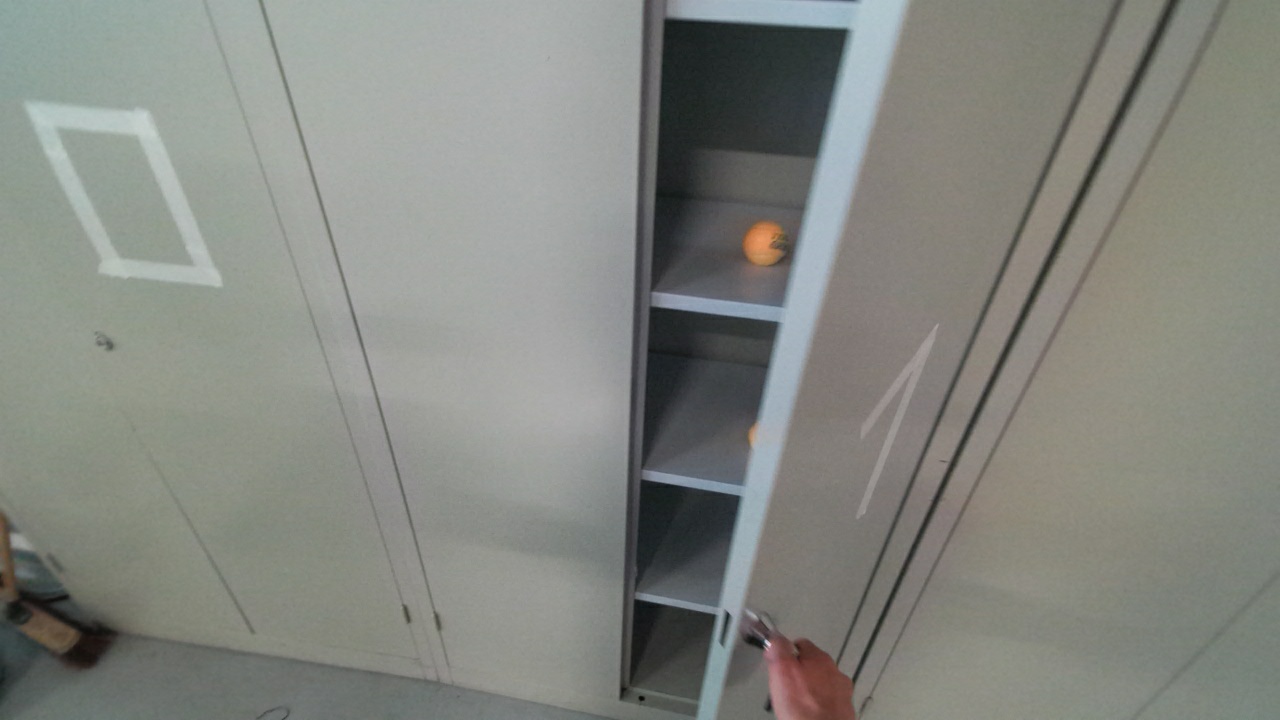}} & 
{\includegraphics[width=0.3\linewidth]{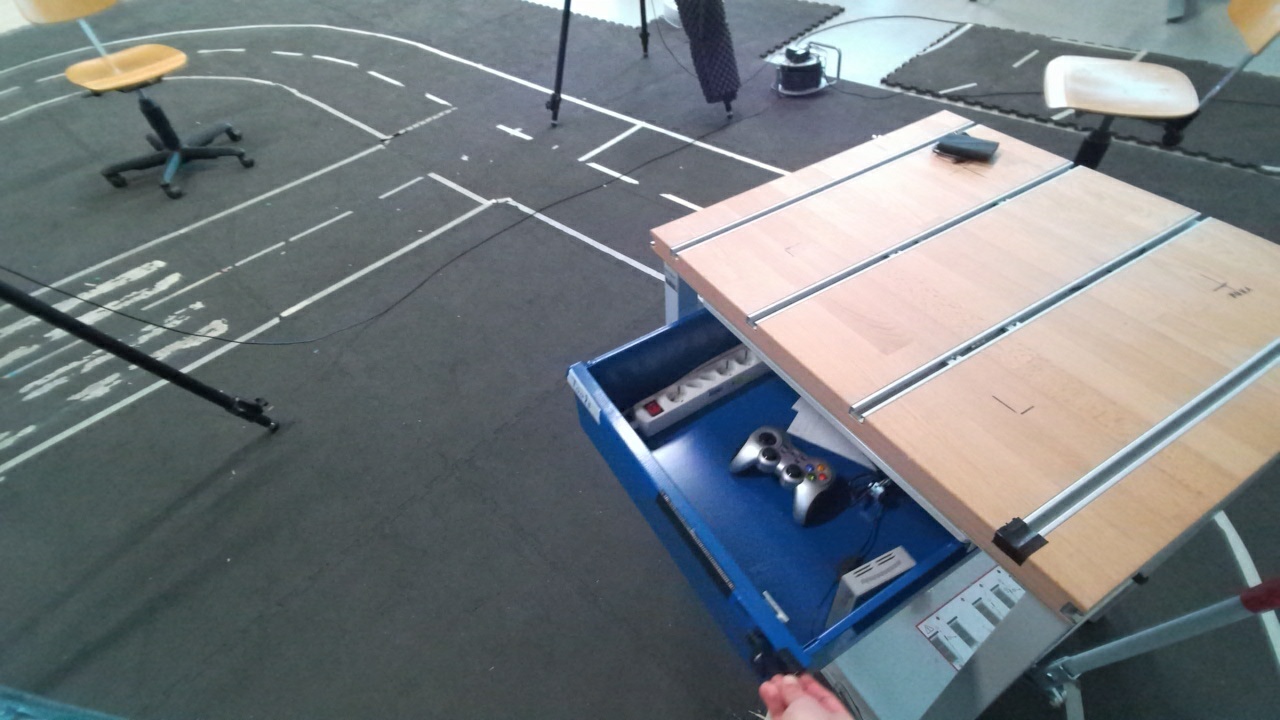}} & 
{\includegraphics[width=0.3\linewidth]{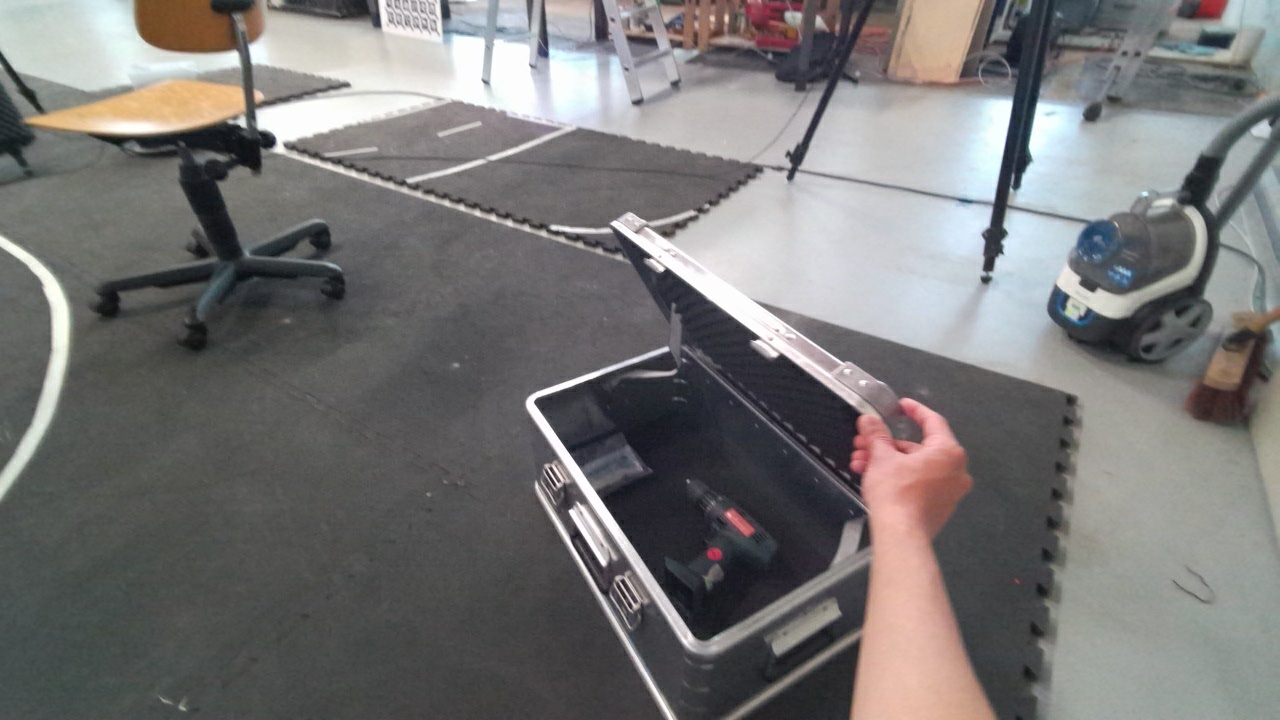}} \\
&
{\includegraphics[width=0.3\linewidth]{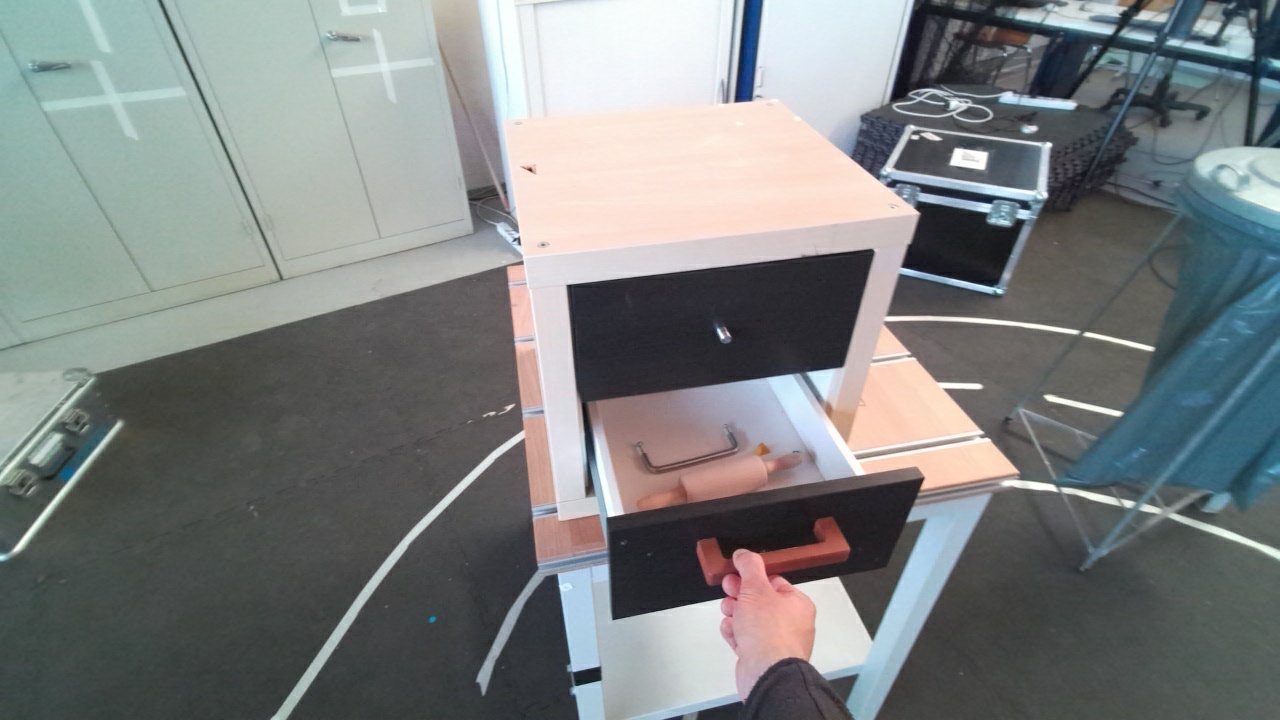}} &
{\includegraphics[width=0.3\linewidth]{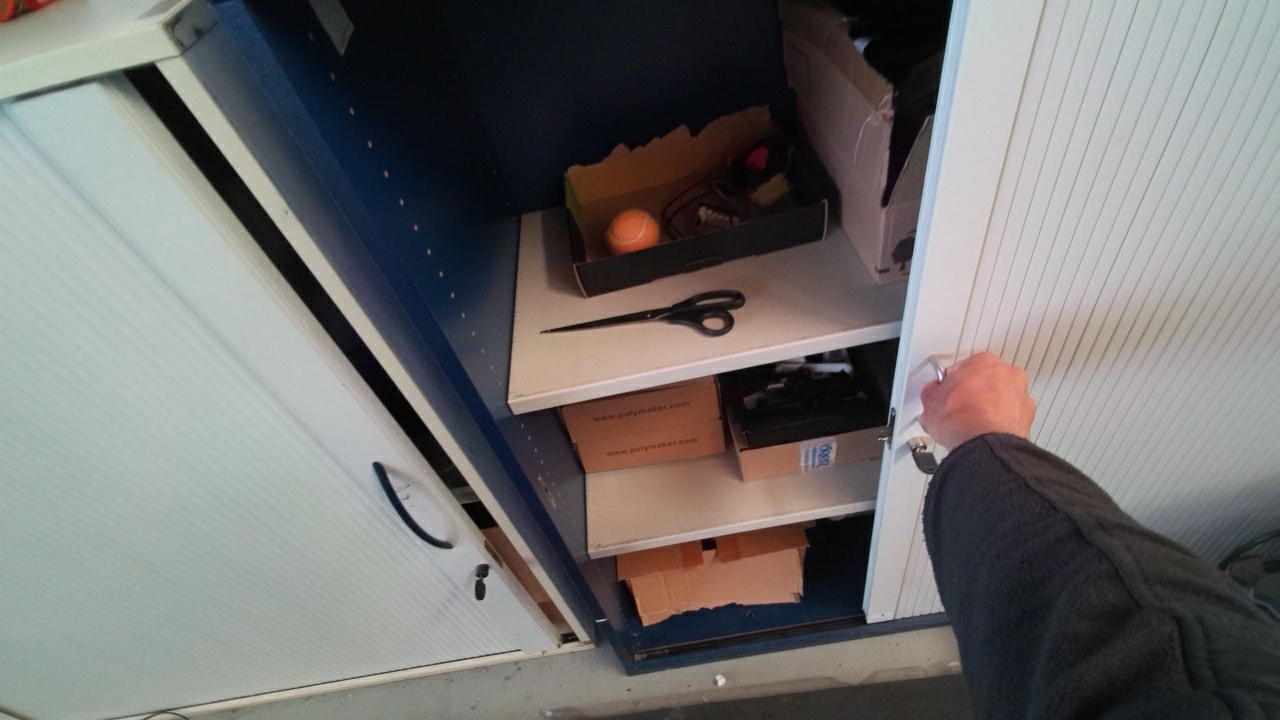}} &
{\includegraphics[width=0.3\linewidth]{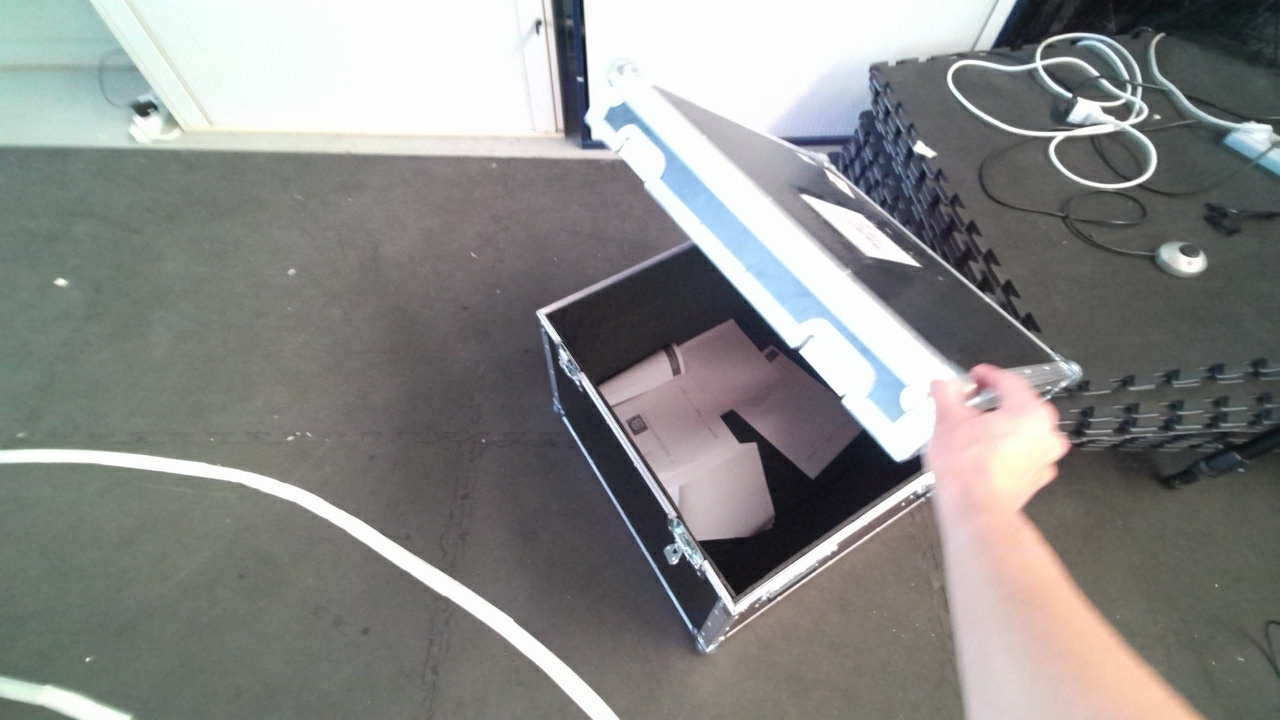}} 
\\
\\
\\

\multirow{2}{*}{\texttt{RH201}} &
{\includegraphics[width=0.3\linewidth]{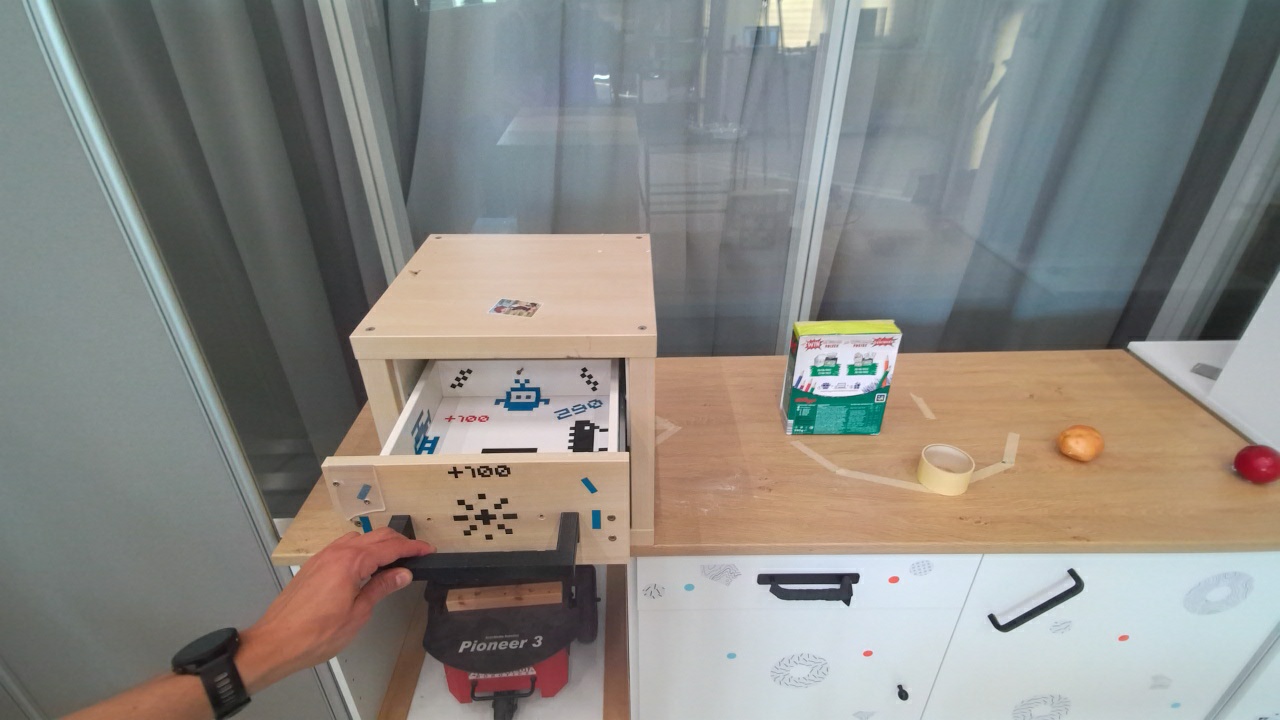}} & 
{\includegraphics[width=0.3\linewidth]{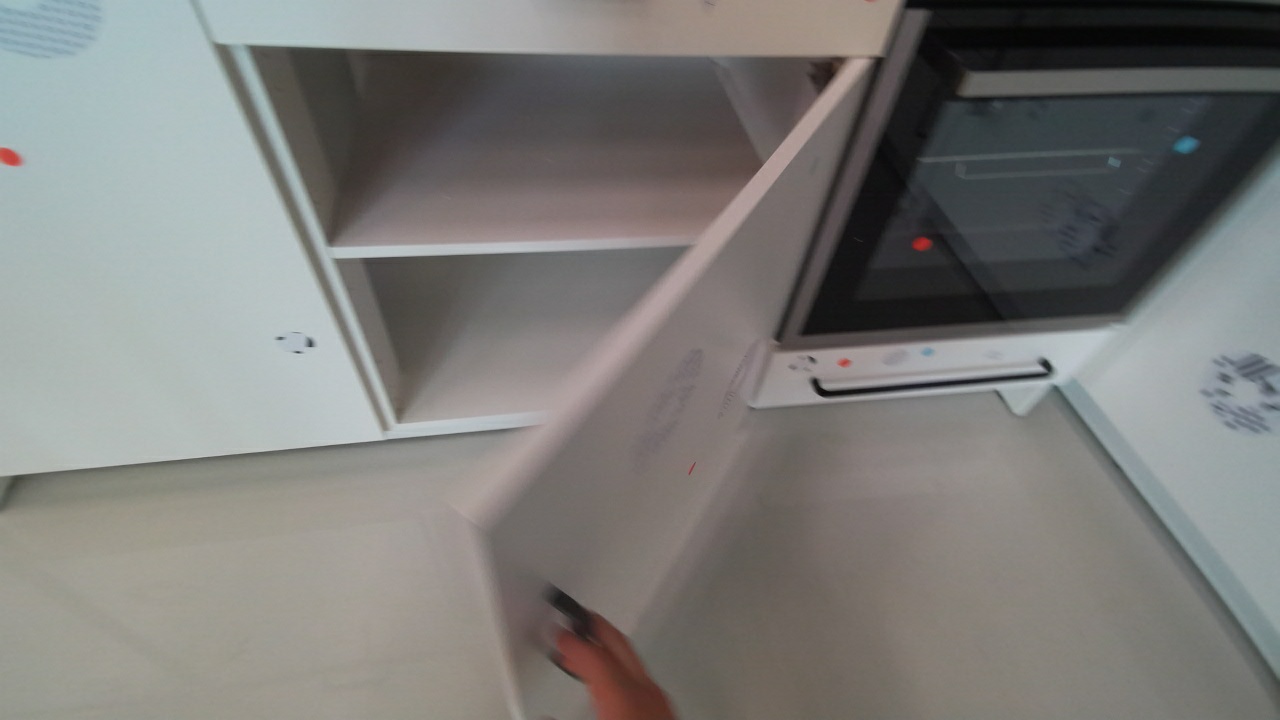}} & 
{\includegraphics[width=0.3\linewidth]{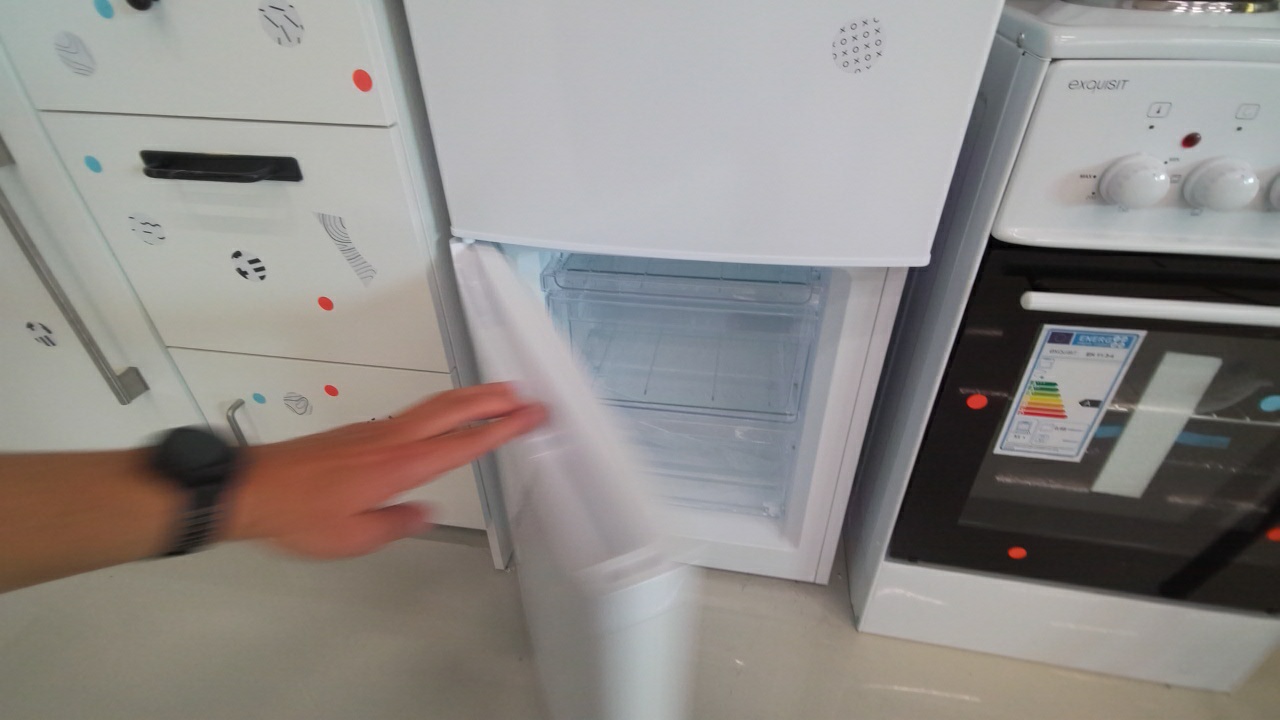}}
\\
&
{\includegraphics[width=0.3\linewidth]{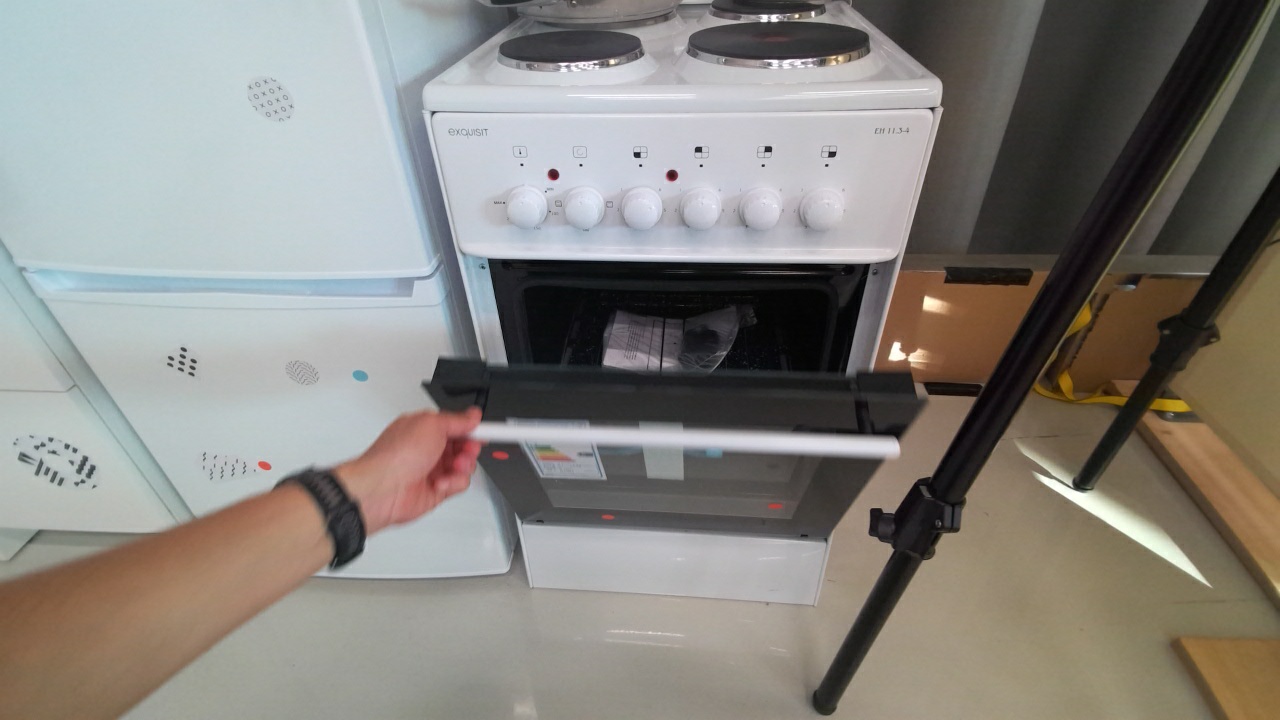}} & 
{\includegraphics[width=0.3\linewidth]{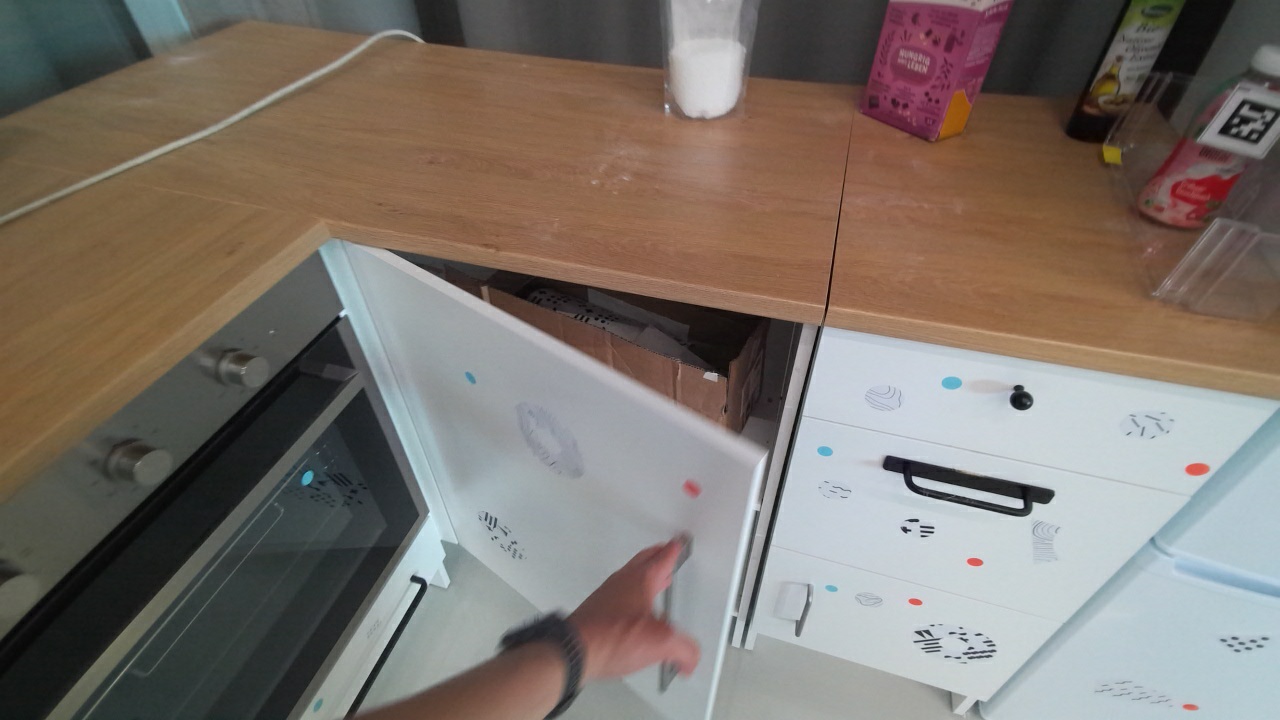}} & 
{\includegraphics[width=0.3\linewidth]{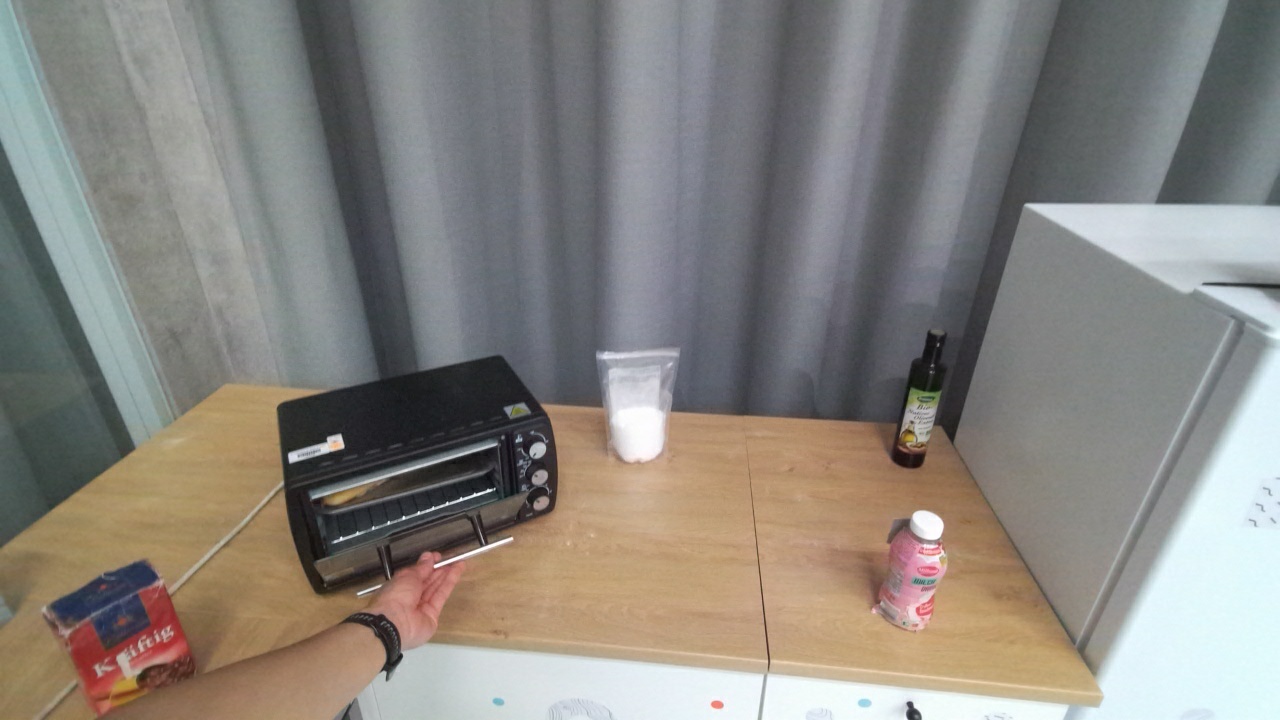}}
\\
\\
\\

\multirow{2}{*}{\texttt{RR080}} &
{\includegraphics[width=0.3\linewidth]{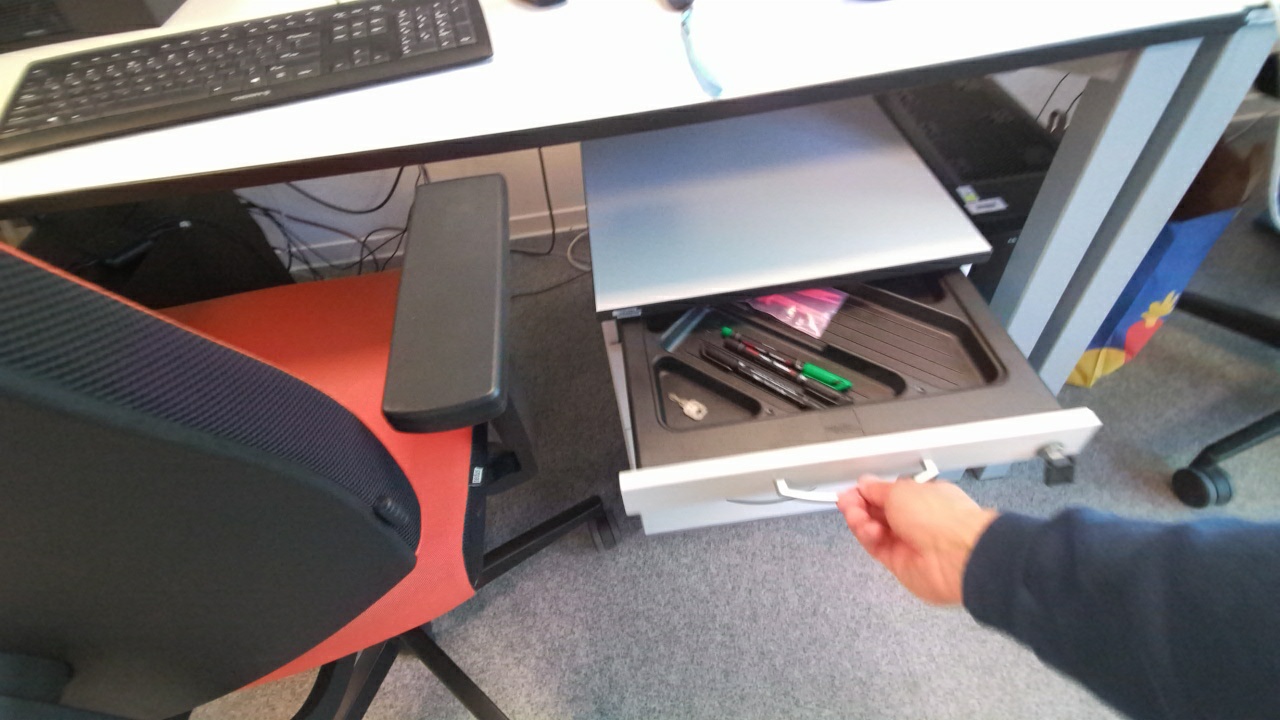}} & 
{\includegraphics[width=0.3\linewidth]{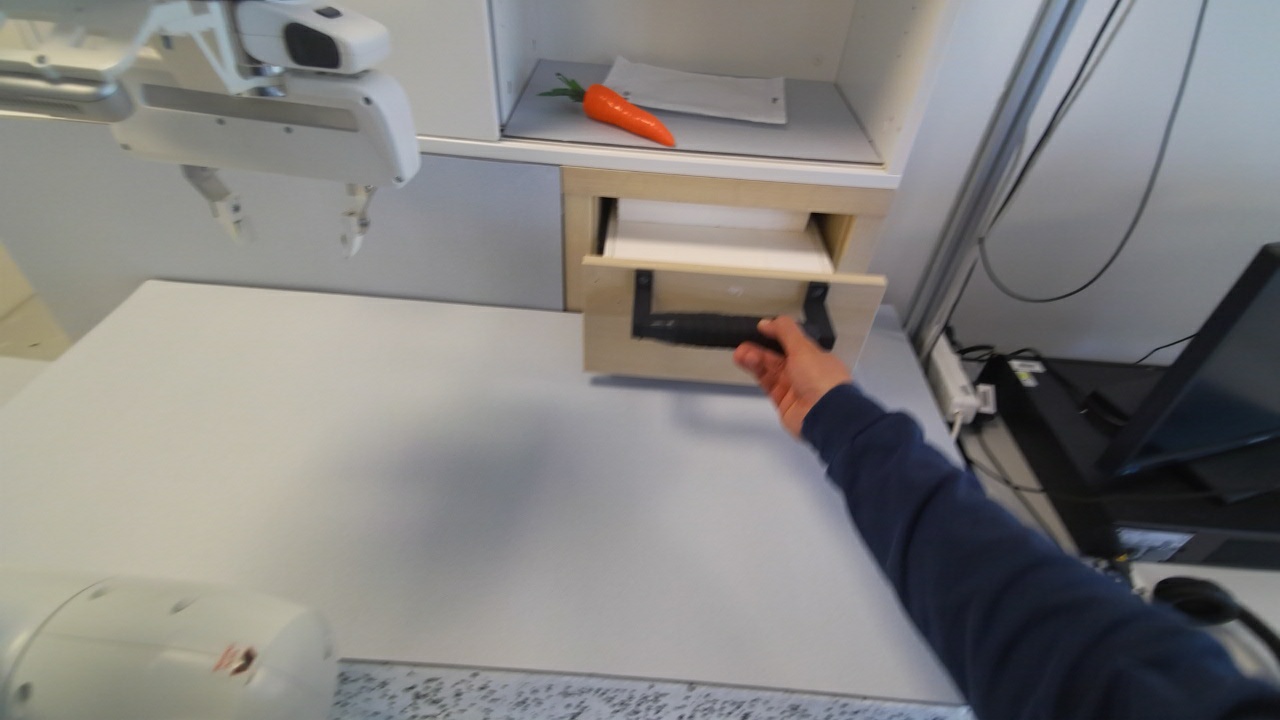}} & 
{\includegraphics[width=0.3\linewidth]{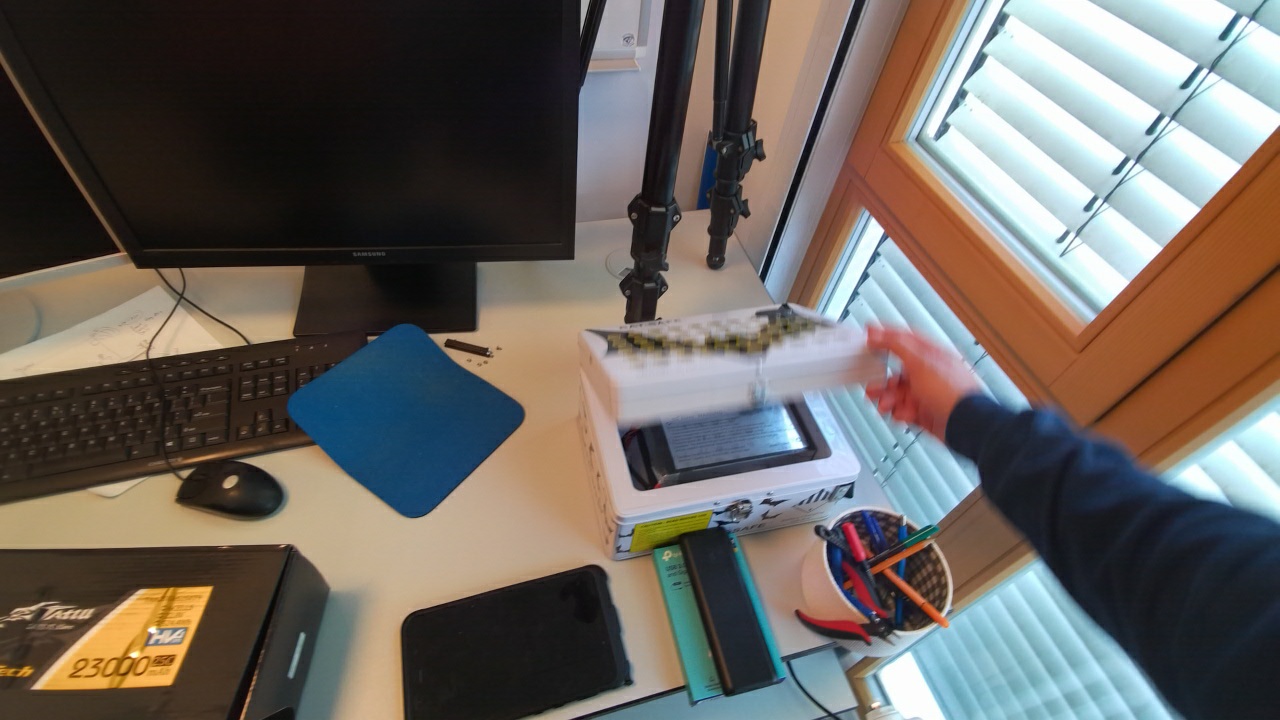}}
\\
 &
{\includegraphics[width=0.3\linewidth]{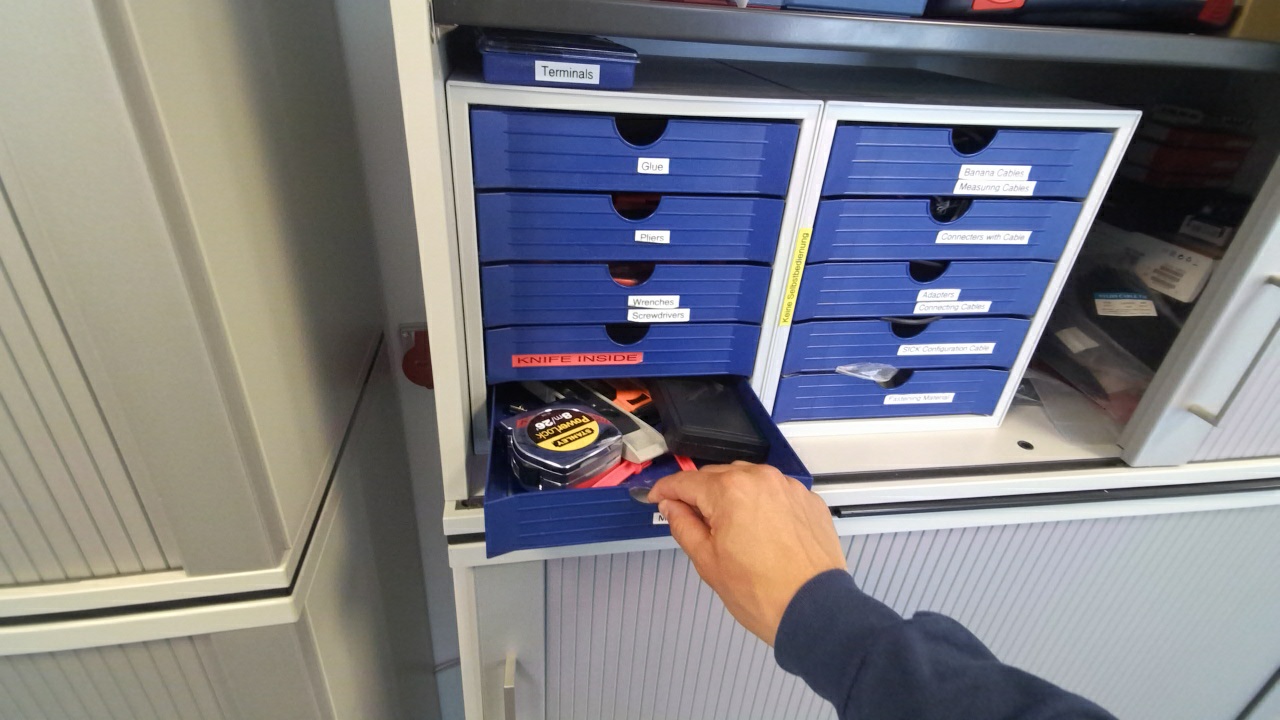}} & 
{\includegraphics[width=0.3\linewidth]{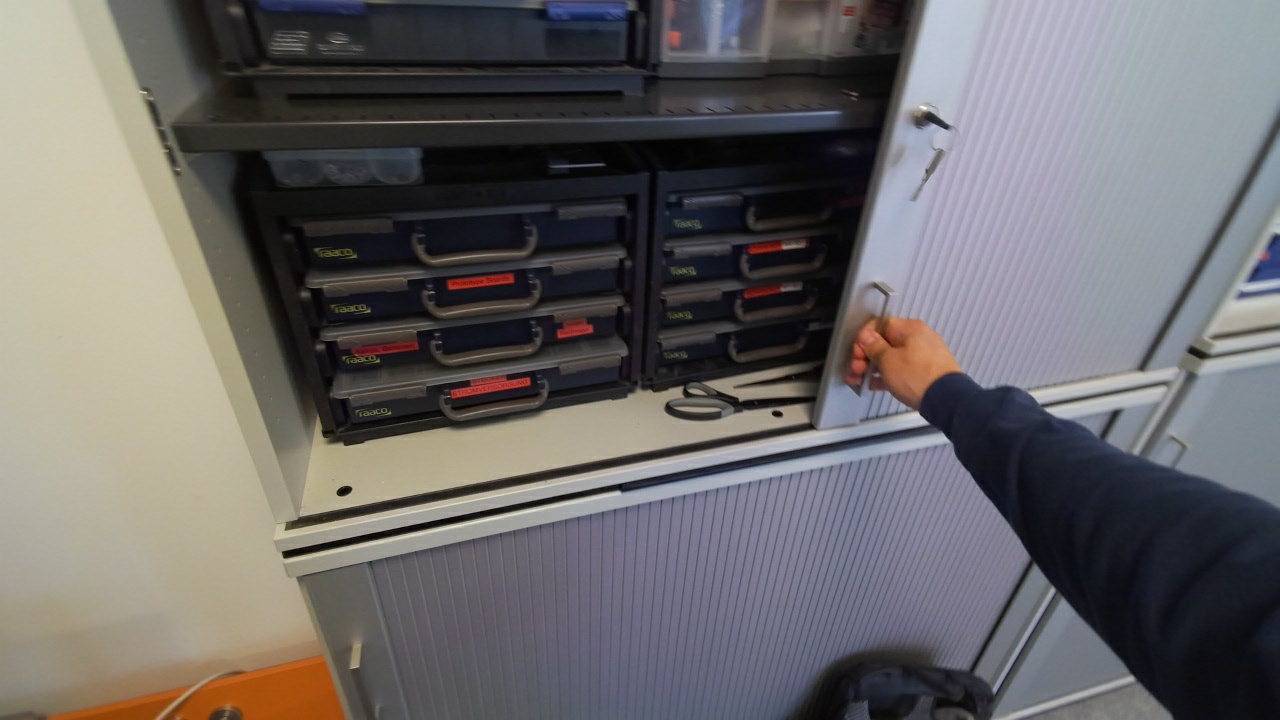}} & 
{\includegraphics[width=0.3\linewidth]{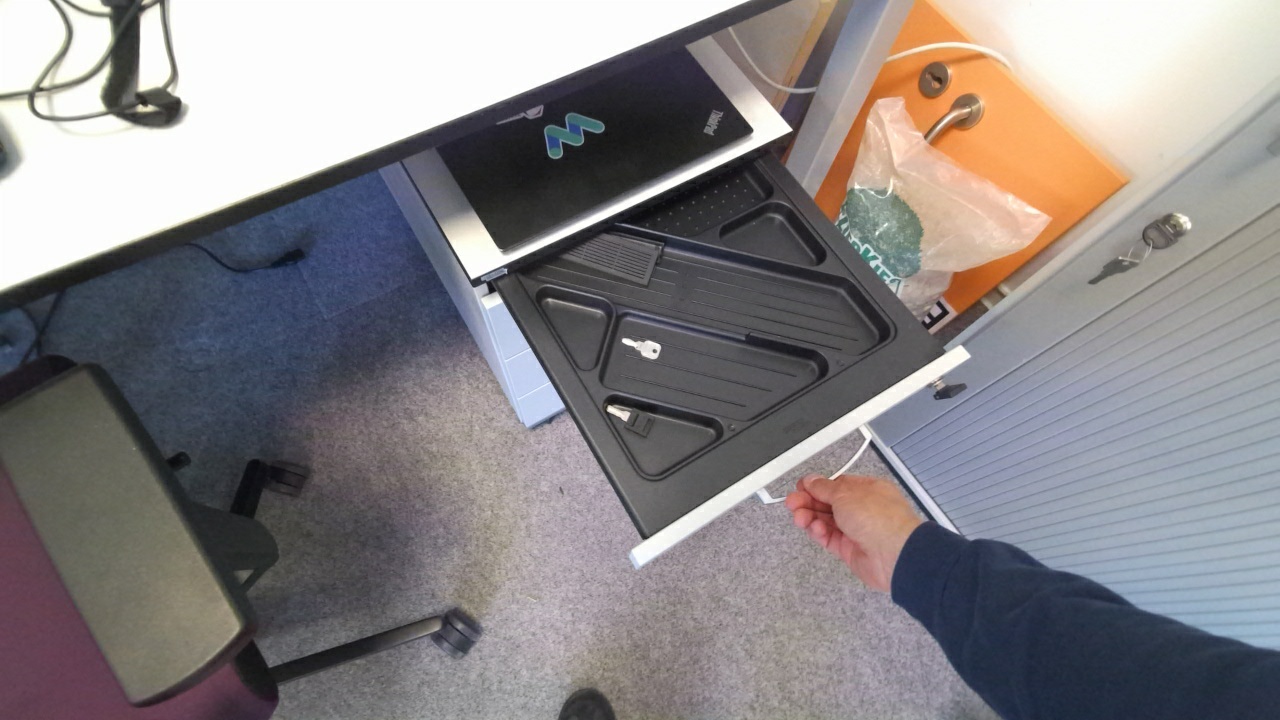}} 
\\
\\

\end{tabular}
}
\caption{We visualize several object interactions across the four different environments (\texttt{DR080}, \texttt{RH078}, \texttt{RH201}, \texttt{RR080}) captured as part of the Arti4D dataset.}
\label{fig:arti4d-examples}

\end{figure*}

{\footnotesize
\bibliographystyleS{IEEEtran}
\bibliographyS{bibliography}
}

\end{document}